\documentclass[10pt,twocolumn,letterpaper]{article}

\usepackage{cvpr}
\usepackage{booktabs} % For formal tables
\usepackage{times}
\usepackage{epsfig}
\usepackage{graphicx}
\usepackage{bm}
\usepackage{amsmath}
\usepackage{amssymb}
\usepackage{tabularx}
\usepackage{microtype}
\usepackage[american]{babel}
\usepackage[detect-all]{siunitx}
\usepackage{etoolbox}
\usepackage{enumitem}
\usepackage{subcaption}
\usepackage{multirow}
\usepackage[skip=2pt]{caption} 
\newrobustcmd*{\bftabnum}{%
  \bfseries
  \sisetup{output-decimal-marker={\textmd{.}}}%
}
\usepackage[breaklinks=true,bookmarks=false]{hyperref}
\hypersetup{
    pdfnewwindow=false,      % links in new PDF window
    colorlinks=true,       % false: boxed links; true: colored links
    linkcolor=red,          % color of internal links (change box color with linkbordercolor)
    citecolor=green,        % color of links to bibliography
    filecolor=magenta,      % color of file links
}

\cvprfinalcopy % *** Uncomment this line for the final submission

\def\cvprPaperID{4419} % *** Enter the CVPR Paper ID here

\begin{document}

%%%%%%%%% TITLE
\title{Complete the Look: Scene-based Complementary Product Recommendation}
% \title{Complete the Look: Scene-based Complementary Recommendation}
%JULIAN: Messed with the title a little. Is "scene image" a common term? Could also be "in the wild" or "real world images" or similar.
%WC: There are many works on "scene understanding", a definition is here, e.g. http://people.csail.mit.edu/torralba/courses/6.870/slides/lecture5.pdf. I think it's a relatively general term, though in our context we just use it to distinguish with product images.
% \title{Complete the Look: Recommending Complements from Scene Images}

\author{Wang-Cheng Kang\textsuperscript{\dag}\thanks{Work done while intern at Pinterest.} , Eric Kim\textsuperscript{\ddag}, Jure Leskovec\textsuperscript{\ddag\S}, Charles Rosenberg\textsuperscript{\ddag}, Julian McAuley\textsuperscript{\dag}\\
\textsuperscript{\ddag}Pinterest, \textsuperscript{\S}Stanford University, \textsuperscript{\dag}UC San Diego\\
{\tt\small \{wckang,jmcauley\}@ucsd.edu, \{erickim,jure,crosenberg\}@pinterest.com}
}


\maketitle
\thispagestyle{empty}

%%%%%%%%% ABSTRACT
\begin{abstract}
Modeling fashion compatibility is challenging due to its complexity and subjectivity. Existing work focuses on predicting compatibility between product images (e.g.~an image containing a t-shirt and an image containing a pair of jeans). However, these approaches ignore real-world `scene' images (e.g.~selfies); 
%JULIAN: Tried to rewrite. Such images can be both good and bad!
%containing key context (e.g., body shapes, seasons, seasons), which limits their performance and applicability.
such images are hard to deal with due to their complexity, clutter, variations in lighting and pose (etc.)~but on the other hand could potentially provide key context (e.g.~the user's body type, or the season) for making more accurate recommendations.
% might improve performance.
In this work, we propose a new task called `Complete the Look', which seeks to recommend visually compatible products based on scene images. 
%JULIAN: Mentioned the dataset construction as one of the contributions. Not sure if this is desirable
We design an approach to extract training data for this task,
%We design 
and propose
a novel way to learn the scene-product compatibility from fashion or interior design images. Our approach measures compatibility both globally and locally via CNNs and attention mechanisms. Extensive experiments show that our method achieves significant performance gains
%JULIAN: Over what?
over alternative systems. 
Human evaluation and qualitative analysis are also conducted to further understand model behavior. We hope this work could lead to useful applications which link 
%massive 
large corpora of
real-world scenes with shoppable products.
\end{abstract}
\vspace{-0.8cm}
%%%%%%%%% BODY TEXT
%erickim: Is it possible to tweak the intro to put more emphasis on how we are learning a "style" space, where scenes and products are jointly represented in the same "style" feature space? I think this is a neat core modeling approach that readers can appreciate. If so, then we should double down on this idea throughout the paper, and repeatedly mention that we're learning a latent style space (or something). To me, learning a "style" feature space sounds more engaging than learning a scene-product similarity function.
\section{Introduction}

% following the instruction from https://cs.stanford.edu/people/widom/paper-writing.html
% 1, What is the problem?
% 2, Why is it interesting and important?  
% 3, Why is it hard? (E.g., why do naive approaches fail?)  data, method
% 4, Why hasn't it been solved before? (Or, what's wrong with previous proposed solutions? How does mine differ?)  product-based
% 5, What are the key components of my approach and results? Also include any specific limitations.

%erickim: perhaps we can reword the first sentence to something like: "Visual signals are a key feature for fashion analysis. By utilizing advances in deep learning for computer vision, both academia~\cite{DBLP:conf/iccv/KiapourHLBB15,DBLP:conf/cvpr/LiuLQWT16} and industry~\cite{DBLP:conf/kdd/YangKBSWKP17,DBLP:conf/kdd/JingLKZXDT15,DBLP:conf/kdd/ZhangPZZZRJ18} have implemented various fashion tasks, ranging from clothing recognition~\cite{DBLP:conf/cvpr/LiuLQWT16,wang2018attentive} to fashion retrieval~\cite{DBLP:conf/iccv/KiapourHLBB15}."
%WC: changed.
Visual signals are a key feature for fashion analysis. 
%By utilizing advances in deep learning for computer vision, 
Recent advances in deep learning 
% for computer vision 
have been adopted by
both academia~\cite{DBLP:conf/iccv/KiapourHLBB15,DBLP:conf/cvpr/LiuLQWT16,wang2018attentive} and industry~\cite{DBLP:conf/kdd/JingLKZXDT15,DBLP:conf/kdd/YangKBSWKP17,DBLP:conf/www/ZhaiKJFTDDD17,DBLP:conf/kdd/ZhangPZZZRJ18} 
%have implemented 
to realize
various fashion-related applications,
%JULIAN: Not much of a range! Doesn't get much more traditional than recognition and retrieval!
ranging from 
clothing recognition to fashion retrieval.
% As visual signals are a key factor for fashion analysis, and benefited by recent advance of deep learning for computer vision, both academia~\cite{DBLP:conf/iccv/KiapourHLBB15,DBLP:conf/cvpr/LiuLQWT16} and industry~\cite{DBLP:conf/kdd/YangKBSWKP17,DBLP:conf/kdd/JingLKZXDT15,DBLP:conf/kdd/ZhangPZZZRJ18} has designed and implemented various deep learning based approaches for different fashion tasks ranging from clothing recognition~\cite{DBLP:conf/cvpr/LiuLQWT16,wang2018attentive} to fashion retrieval~\cite{DBLP:conf/iccv/KiapourHLBB15}.
Fashion images can be categorized into scene images (fashion images in the wild) and product images (fashion item images from shopping websites, usually containing a single product on a plain background). 
Generally speaking, 
%scene photos 
the former
(e.g.~selfies, street photos) predominate on image sharing applications, 
%while product images 
whereas the latter
are more common on online shopping websites. 
%WC: not sure if it's suitable to mention the above sentence here.
% To link the two kinds of fashion images, a cross-scenario fashion retrieval tasks (called street2shop) has been proposed, which seeks to learn the notion of similarity between products in real world images and shop photos~\cite{DBLP:conf/iccv/KiapourHLBB15,DBLP:conf/cvpr/LiuSLXLY12}. 
%erickim: we should reword this so that it doesn't sound like the task is learning the complementarity, something like: To link scene and product images, we propose a new task called \emph{Complete the Look}, the aim of which is to learn scene-to-product complements, such that we can recommend products that go well with a given scene.

\begin{figure}[t]
\centering
\includegraphics[width=\linewidth]{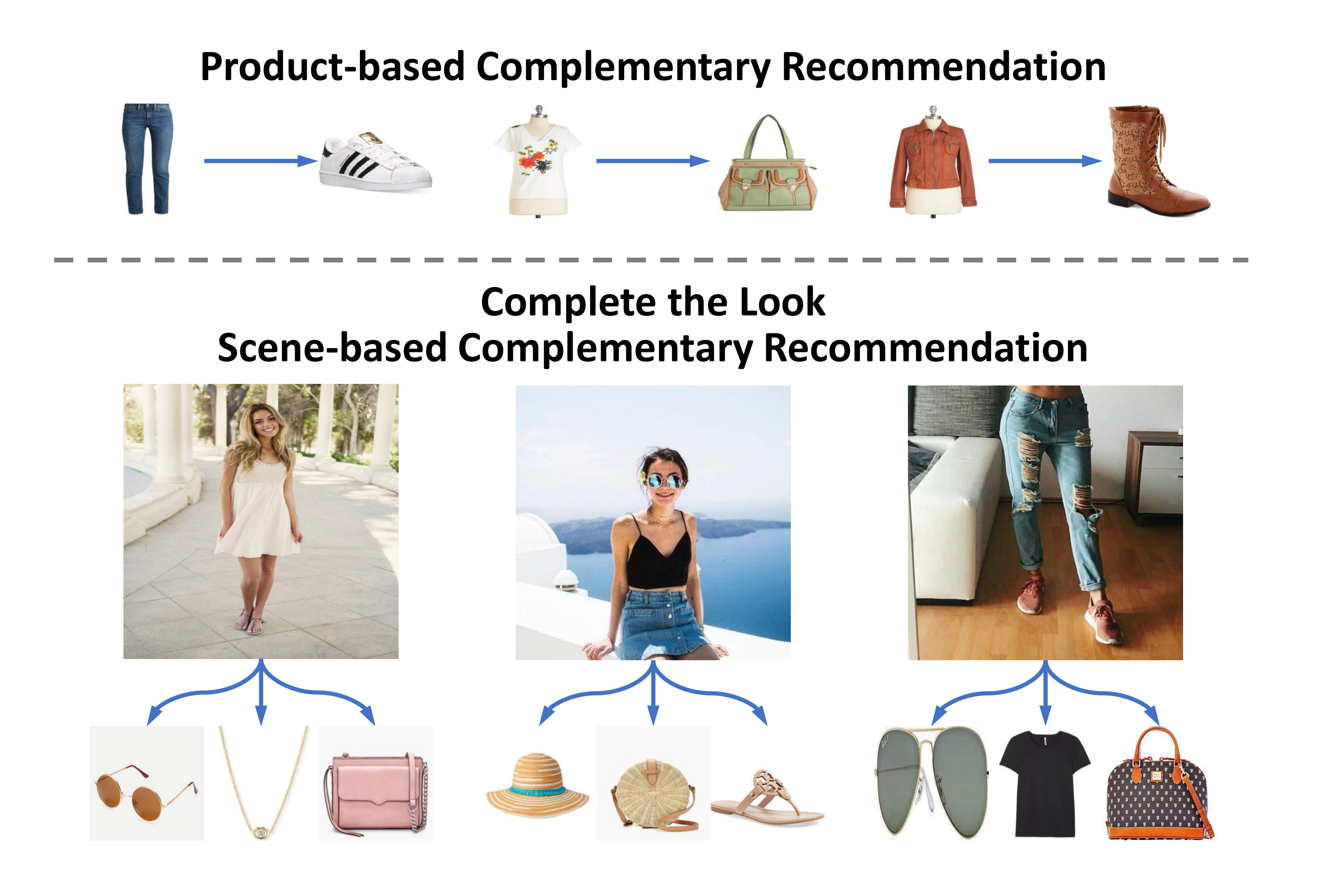}
\caption{A comparison of the use cases of product-based and scene-based complementary product recommendation. Our approach (bottom) seeks to recommend compatible fashion items based on a real-world scene, while product-based approaches (top) consider compatibility between products.}
\label{fig:intro}
\end{figure}

%JULIAN: Promoted to its own paragraph, but could lead into the next one
%To link the 
We seek to bridge the gap between these
%two kinds 
two types
of images,
% * <erickim555@gmail.com> 2018-11-16T19:49:41.928Z:
% 
% > image
% images
% 
% ^ <erickim555@gmail.com> 2018-11-16T19:49:52.408Z.
%we propose 
via
a new task called \emph{Complete the Look} (CTL), in which we seek to recommend fashion products from various categories that complement (or `go well with') the given scene (Figure~\ref{fig:intro}).
%JULIAN: Added some description of why this setting is relevant
Compared to existing approaches, this setting corresponds more closely to real-world use-cases in which users might seek recommendations of complementary items, based on images they upload `in the wild.'

%erickim: do you want to add a final small paragraph that summarizes the contributions of this paper? something like (roughly): "The contributions of this paper are as follows: (1) A new fashion task, Complete the Look (2) A novel model architecture that captures scene-product compatibility, achieving state-of-the-art results on fashion task tasks."
%WC: Yep, I assume we will state the contribution at the last paragraph of the intro.

%erickim: I really like this Figure 1! Great work.

%erickim: to me, it seems that this paragraph can go in the related works section.
Fashion compatibility has been studied previously~\cite{DBLP:conf/iccv/VeitKBMBB15,DBLP:conf/mm/HanWJD17}, 
%and existing approaches 
though existing approaches
mainly consider only product images (Figure \ref{fig:intro}). In comparison, our scene-based CTL task has three significant features: 1)~scene images contain not only the fashion items worn by the subject (or user), but also rich context like their body type, the season, etc. By exploiting this side-information, we can potentially provide more accurate and customized recommendations; 
2)~our system can be adopted by users to give fashion advice (e.g.~shoes that go well with your outfit) 
simply by uploading (e.g.)~a selfie;
%for themselves via uploading a selfie;
% 2)~the task closely corresponds to real-world scenarios, e.g.~a user uploading a selfie;
%JULIAN: Does this matter? It's quite vague what you mean by "organically linked" and it's not really a research contribution. Without revealing who you are it's hard to even say what you really mean. I tried to edit.
%WC: I assume this paragraph is to sell the idea of the proposed 'task' to show it's useful and important (especially against product-based) rather than contribution? 2)&3) are trying to state two use cases for single user and website
3)~our system can be readily adapted to existing platforms to recommend products appearing in fashion images.
% 3)~we are able to leverage both public and proprietary data to build large training sets consisting of scene images and their corresponding shoppable products.
%WC: though we won't release the data, it's not necessary to point out it's proprietary.

Learning scene-product compatibility is at the core of the CTL task. However, 
%getting 
constructing
appropriate 
ground-truth data to learn the notion of compatibility 
%is one 
%of the challenges for the problem. 
%of the main challenges we face.
is a significant challenge.
Existing large-scale fashion datasets are 
%JULIAN: Avoid this word, it has a too specific meaning (i.e., > 50%) which is not something you can back.
%mainly 
typically
labeled with clothing segments, attributes, or landmarks~\cite{DBLP:conf/cvpr/LiuLQWT16,DBLP:conf/wacv/AkLTK18,DBLP:conf/mm/ZhengYKP18}, 
%where 
%compatibility information is absent. 
%erickim[2018-11-16] suggestion to reword this phrase to: "Existing large-scale fashion datasets..., or landmarks, which are absent OF any..."
which are absent of any information regarding compatibility.
Product-based methods have adopted (for example) Amazon's co-occurrence data~\cite{DBLP:conf/iccv/VeitKBMBB15,DBLP:conf/sigir/McAuleyTSH15} or Polyvore's outfit data~\cite{DBLP:conf/mm/HanWJD17,DBLP:conf/eccv/VasilevaPDRKF18,DBLP:conf/mm/SongFLLNM17} to learn product-to-product compatibility. However, these datasets can not be used for our CTL task, as they 
%only contain product images. 
lack images from real-world scenes.
In addition to the problem of data 
availability, another challenge is to estimate the compatibility between product images and real-world fashion images,
%of which the characteristics significantly differ from each other.
whose characteristics can differ significantly.

%erickim[2018-11-16] This paragraph feels like it belongs in the Related Works section.
As mentioned above, existing 
%fashion compatibility 
studies typically consider
%product compatibility~
compatibility of product images~\cite{DBLP:conf/iccv/VeitKBMBB15,DBLP:conf/mm/HanWJD17}, 
%and new datasets and 
meaning that new data and
techniques must be introduced for our CTL task. Another line of related work considers a cross-scenario fashion retrieval task called Street2Shop 
%(aka 
(also known as
\emph{Shop the Look}, or STL)~\cite{DBLP:conf/iccv/KiapourHLBB15,DBLP:conf/cvpr/LiuSLXLY12} which seeks to retrieve similar-looking (or even identical) products given a scene image and a bounding box of the query product. 
% To estimate cross-scenario similarity, 
Human-labeled datasets have been introduced to estimate cross-scenario similarity~\cite{DBLP:conf/iccv/KiapourHLBB15}, though our CTL task differs from STL in that we seek to learn a notion of complementarity (instead of similarity), and critically the desired complementary products typically \emph{don't} appear in the given scene (Figure~\ref{fig:data}).

%erickim[2018-11-16] On my first reading of the first sentence of this paragraph, I got the impression that the main contribution of this paper is the "approach to generating CTL datasets based on STL data via cropping". One way to avoid this confusion is to start with: "The contributions of this paper are as follows. We design an approach to generate CTL datasets based on STL data via cropping. We learn global embeddings from scene and product images...", ie listing out the paper contributions. But if you think it's OK as-is, then feel free to leave it.
In this paper, we design an approach to generate CTL datasets based on STL data via cropping. In addition to 
%using datasets in the fashion domain as other works, 
leveraging existing datasets from the fashion domain,
we also consider 
%a related yet different domain, interior design. 
the domain of interior design (Section \ref{sec:data}).
We learn global embeddings from scene and product images and local embeddings from local regions of the scenes, and measure scene-product compatibility in a unified style space with category-guided attention~(Section \ref{sec:method}). We evaluate both the overall and Top-K ranking performance of our method against various baselines, quantitatively and qualitatively 
analyze
the attended 
regions, and perform a user study to measure the difficulty of the task~(Section \ref{sec:exps}). 
%areas. 
% Extensive results show the 
% %JULIAN: Again, would reword. You're not really "better" than the other baselines, they were mostly just not designed for this task.
% superiority 
% of our method against various 
% %JULIAN: Sentence borked, I assume you're still writing
% baselines. and qualitative results

\section{Related Work}

\textbf{Visual Fashion Understanding.} Recently, computer vision for fashion has attracted significant attention, 
%and shown its power and potential in various applications especially with 
with various applications typically built on top of
deep convolutional networks. Clothing `parsing' is one such application, which seeks to parse and categorize 
%all the 
garments in a fashion image~\cite{DBLP:conf/cvpr/YamaguchiKOB12,DBLP:journals/tmm/LiangLYLHY16,DBLP:journals/pami/YamaguchiKOB15}. Since clothing has fine-graind style attributes (e.g.~sleeve length, material, etc.), some 
%work seeks 
works seek
to 
identify clothing attributes~\cite{DBLP:conf/accv/BossardDLWQG12,DBLP:conf/eccv/KiapourYBB14,DBLP:conf/iccv/Al-HalahSG17}, and detect fashion landmarks (e.g.~sleeve, collar, etc.)~\cite{wang2018attentive,DBLP:conf/cvpr/LiuLQWT16}. Another line of work considers retrieving fashion images based on various forms of queries, including images~\cite{DBLP:conf/cvpr/LiuLQWT16, DBLP:conf/cvpr/Simo-SerraI16,DBLP:conf/www/ZhaiKJFTDDD17},  attributes~\cite{di2013style,ak2018learning}, occasions~\cite{DBLP:conf/mm/LiuFSZLXY12}, videos~\cite{DBLP:conf/cvpr/ChengWLH17}, and user preferences~\cite{DBLP:conf/icdm/KangFWM17}. Our work is closer to the `cross-scenario' fashion retrieval setting (called street2shop) which seeks to retrieve fashion products appearing in street photos~\cite{DBLP:conf/cvpr/LiuSLXLY12, DBLP:conf/iccv/KiapourHLBB15}, 
%as our datasets can be converted from theirs.
as the same type of data can be adapted to our setting.

\textbf{Complementary Item Recommendation.}
Some recent works seek to identify whether two products are complementary, such that we can recommend complementary products based 
%on what the user already has
on the user's previous purchasing or browsing patterns~\cite{DBLP:conf/kdd/McAuleyPL15,DBLP:conf/recsys/ZhangLNC18,DBLP:conf/wsdm/WangJRTY18}. 
%Similarly, in the fashion domain, it's also useful to measure the visual compatibility between fashion items. 
In the fashion domain, visual features can be useful to determine compatibility between items, for example in terms of pairwise
%Existing work seeks to model pairwise fashion 
compatibility~\cite{DBLP:conf/iccv/VeitKBMBB15,DBLP:conf/mm/SongFLLNM17,DBLP:conf/eccv/VasilevaPDRKF18,DBLP:conf/sigir/McAuleyTSH15}, or outfit compatibility~\cite{DBLP:journals/tmm/LiCZL17,DBLP:conf/mm/HanWJD17,hsiao2018creating,DBLP:conf/wacv/TangsengYO18}. The former
%one 
setting
takes a fashion item as a query and seeks to recommend compatible items from different categories (e.g.~recommend jeans given a t-shirt). The latter seeks to select fashion items to form compatible outfits, or to complete a partial outfit. Our method retrieves compatible products based on a real-world scene containing rich context (e.g.~garments, body shapes, occasions), which can also be viewed as a form of complementary recommendation. However this differs from existing methods 
%as they 
which
%only consider product images and seek to model product-product 
seek to model product-product 
compatibility from pairs of images containing products.
In addition to retrieving existing products, one recent approach uses generative models to generate compatible fashion items~\cite{DBLP:conf/aaai/ShihCLS18}.

% Deep Similarity:

\textbf{Attention Mechanisms.}
`Attention' has been widely used in computer vision tasks including image captioning~\cite{DBLP:conf/icml/XuBKCCSZB15,DBLP:conf/cvpr/ChenZXNSLC17}, visual question answering~\cite{DBLP:conf/cvpr/ShihSH16,DBLP:conf/eccv/XuS16}, image recognition~\cite{DBLP:conf/nips/JaderbergSZK15,DBLP:conf/cvpr/WangJQYLZWT17}, and generation~\cite{xu2017attngan,zhang2018self}. 
%The attention 
Attention
is mainly used to `focus' on relevant regions of an image (known as `spatial attention'). To identify relevant regions of fashion images, previous methods have adopted pretrained 
%human 
person
detectors to segment images~\cite{DBLP:conf/cvpr/LiuSLXLY12,song2011predicting}. Another approach discovers relevant regions by 
attribute activation maps (AAMs)~\cite{DBLP:conf/cvpr/ZhouKLOT16}, generated using labels including clothing attributes~\cite{ak2018learning} and descriptions~\cite{DBLP:conf/iccv/HanWHZZLZD17}. Recently, attention mechanisms have 
%been adopted and 
achieved strong performance on visual fashion understanding tasks like clothing 
%category classification 
categorization
and fashion landmark detection~\cite{wang2018attentive}. Our work is the first (to our knowledge)
%applying the 
to apply
attention to discover relevant regions guided by supervision in the form of compatibility.
%JULIAN: Last bit necessary? This text is a bit too specific when describing what makes the work original -- the more qualifiers you have, the less original it seems. Might be better to drop in favor of explicitly stating the original aspects of the contribution in the intro.
%WC: I see, removed, let's state it in the intro.
% and to study its effect for interior design.

\textbf{Deep Similarity Learning.} 
% Learning similarity models with deep neural networks has become popular in recent years.
A variety of methods have been proposed to measure similarity with deep neural networks.
Siamese Networks are a classic approach, which seek to learn an embedding space such that similar images have short distances, and have been applied to face verification and dimensionality reduction~\cite{DBLP:conf/cvpr/ChopraHL05, DBLP:conf/cvpr/HadsellCL06}. Recent methods tend to use triplet losses%
%like hinge loss~
~\cite{DBLP:conf/cvpr/SchroffKP15,DBLP:conf/cvpr/WangSLRWPCW14} by considering an anchor image, a positive image (similar to the anchor), and a negative image (randomly sampled), such that the distance from the anchor to the positive image should be less than that of the negative. Recent studies have found that better sampling strategies (e.g.~sampling `hard' negatives) can aid performance~\cite{DBLP:conf/cvpr/SchroffKP15,DBLP:conf/iccv/ManmathaWSK17}. In our method, we seek to learn a unified style space where compatible scenes and products are close, as they ought to represent similar styles.

\section{Datasets}\label{sec:data}

We first introduce datasets for the \emph{Shop the Look} (STL) task, before describing how to 
%appropriately 
convert STL data into a format that can be used for our \emph{Complete the Look} (CTL) task.

\subsection{Shop the Look Datasets}

\begin{figure}[t]
\centering
\includegraphics[width=\linewidth]{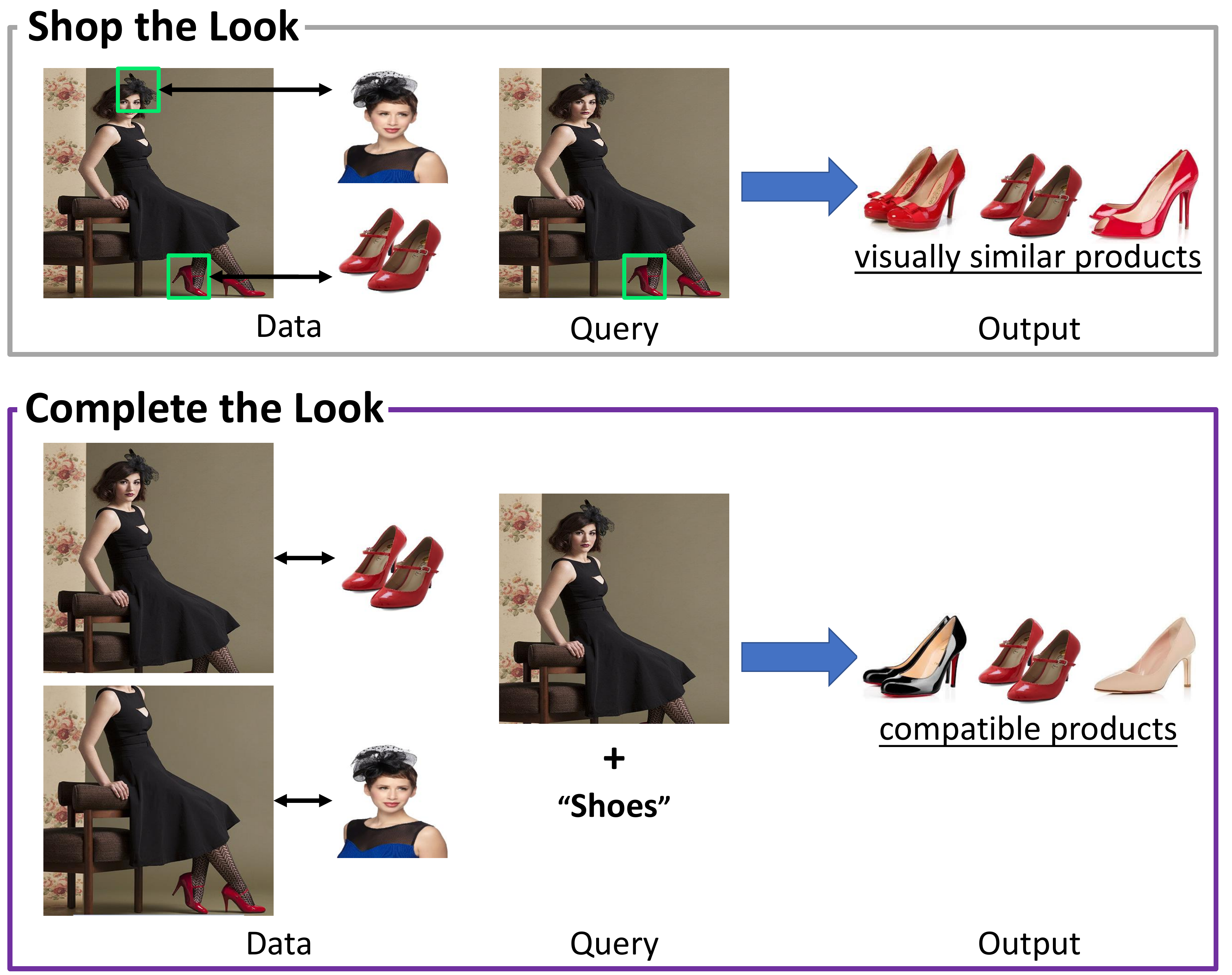}
\caption{A comparison of data formats and tasks between STL and CTL. STL focuses on retrieving similar products while CTL seeks to recommend compatible items that don't appear in the scene. The CTL data contains compatible scene-product (i.e., $I_s$ and $I_p$) pairs.}
\label{fig:data}
\end{figure}

%erickim: The "Street2Shop" task is very similar to the "Complete the Look" task - somewhere in the paper, likely in the Related Works, we should clarify the difference.
As shown in Figure~\ref{fig:data}, the \emph{Shop the Look} (aka Street2Shop) task consists of retrieving visually similar (or even identical) products based on a scene image and a bounding box containing the query product. This application is useful, for example, when a user sees a celebrity wearing 
%a fancy purse, they can 
an item (e.g.~a purse), allowing them to
easily search and buy the product (or a similar one) by taking a photo and 
%draw 
selecting
a bounding box of the item.

The main challenge here arises due to the difference between products (e.g.~clothing) in real-world scenes versus that of online shopping images, where the latter are typically in a canonical pose, on a plain background, adequately lit, etc. To tackle the problem, a recent study sought to collect human-labeled datasets which include bounding boxes of products in scene images, the associated product images, as well as the category of each product~\cite{DBLP:conf/iccv/KiapourHLBB15} (Figure \ref{fig:data}). We describe three datasets that can be used for the \emph{Shop the Look} task as follows:

\textbf{Exact Street2Shop}\footnote{\url{http://www.tamaraberg.com/street2shop/}} Kiapour \emph{et al.}~introduced a first human-labeled dataset for the street2shop task~\cite{DBLP:conf/iccv/KiapourHLBB15}. They crawled data from \emph{ModCloth}, an online fashion store where people can upload photos of themselves wearing 
%ModCloth 
%clothing 
products, 
%and indicate 
indicating
the exact items they
are wearing. \emph{ModCloth} also provides category information for all products. However, since 
%the 
bounding boxes are not provided,
%by users and ModCloth, 
the authors used Amazon's Mechanical Turk to label the bounding box of products in scene images. 

\textbf{Pinterest's Shop The Look\footnote{\url{https://github.com/kang205/STL-Dataset}}} We 
%JULIAN: "Consider" makes it ambiguous whether this is an original dataset or whether you're the first to use it. I'm not sure of the exact provenance of this data so please check
obtained
%consider 
two STL datasets from \emph{Pinterest},
containing various scene images and shoppable products from partners. STL-Fashion contains fashion images and products, while STL-Home includes interior design and home decor items. Both datasets have scene-product pairs, bounding boxes for products, and product category information, all of which are labeled by internal workers. Unlike the Exact Street2Shop dataset~\cite{DBLP:conf/iccv/KiapourHLBB15} 
%that the 
where
users only provide 
%exactly matched products, 
product matches,
here
%JULIAN: I assume you mean internal and not "intern"
%intern
workers also label products that have a \emph{similar} style to the 
%appeared 
observed
product and are compatible with the scene. Furthermore, the two datasets are much larger in terms of both the number of images and scene-product pairs
%JULIAN: added
(Table \ref{tab:data}).

% \subsection{Using STL Data for CTL?}
\subsection{Can STL Data be Used for CTL?}\label{sec:issue}

%Learning the 
Estimating
scene-product compatibility is at the core of the CTL task. Although existing STL datasets provide 
%plenty of 
abundant
scene-product pairs, directly 
%learning compatibility from them 
using them to learn a notion of compatibility
(i.e., viewing each pair as a compatible scene and product) is flawed. 

For example, suppose we 
%learn a 
wanted to learn a
CTL model based on STL data. 
%The model is 
The model might be
trained to predict a high compatibility score for each 
%pair of scene $s$ and product $p^\text{+}$ 
scene/product pair ($s$ / $p^\text{+}$) 
in the STL data, and predict a low compatibility score for the negative product $p^\text{-}$ (e.g.~via random sampling). 
%However, 
That is,
the product $p^\text{+}$ appears in the scene $s$ while the product $p^\text{-}$ doesn't. 
%Hence, it's possibly the model is learned 
Here it is possible that the model will merely learn
to \emph{detect} whether the product appears in the scene (i.e., give a high compatibility score if it appears, and a low score otherwise), instead of 
%to measure the 
measuring
compatibility. 
%If this is the case, 
In this case,
the model would fail on the CTL task which seeks to recommend compatible products which 
%doesn't 
\emph{don't}
appear in the scene. Empirical results also 
%show that this would lead to inferior performance.
show that such an approach leads to inferior performance.
%JULIAN: Argument is okay, but not entirely convincing. E.g. wouldn't a contrastive / ranking loss (as is often used for training with one-class feedback) take care of this issue?

%JULIAN: Also I couldn't spot which method in the experiments corresponds specifically to this idea?

The above %issue is 
issue arises
mainly because the model `sees' the product in the scene image. To address it, we propose a strategy to adapt STL datasets for the CTL task.
%datasets for CTL based on STL data. 
The core idea is to remove the product by cropping the scene image, which forces the model to learn the compatibility between the 
%rest 
remaining
part of the scene image and the product.
%We describe details in the following section.

\begin{table*}
\centering
\footnotesize
\begin{tabularx}{\textwidth}{llrrrX}
\toprule
Name      							& Source		& \#Scene              & \#Product            & \#Pair				& Product Categories (in descending order of quantity)                    \\
\midrule
\textbf{Fashion-1}					&   Exact Street2Shop~\cite{DBLP:conf/iccv/KiapourHLBB15}   			& 10,482               & 5,238                & 10,608              & footwear, tops, outerwear, skirts, leggings, bags, pants, hats, belts, eyewear                                                \\
\multirow{2}{*}{\textbf{Fashion-2}} 					&		\multirow{2}{*}{STL-Fashion (Pinterest)}												& \multirow{2}{*}{47,739}               & \multirow{2}{*}{38,111}               & \multirow{2}{*}{72,198}              & shoes, tops, pants, handbags, coats, sunglasses, shorts, skirts, earrings, necklaces		\\
\multirow{2}{*}{\textbf{Home}} 	    				&		\multirow{2}{*}{STL-Home (Pinterest)}												& \multirow{2}{*}{24,022}               & \multirow{2}{*}{41,306}               & \multirow{2}{*}{93,274}              & rugs, chairs, light fixtures, pillows, mirrors, faucets, lamps, sofas, tables, decor, curtains \\
\bottomrule
\end{tabularx}
% \caption{Data statistics (after preprocessing)}
\caption{Data statistics (after preprocessing). Each pair contains a compatible scene and product.}
%JULIAN: Slightly confusing -- would be better to separate "Task" into "Query" and "Output"
\label{tab:data}
% \vspace{-0.2cm}
\end{table*}

%erickim: To me, this section ("Generating CTL Datasets") belongs in the training methodology section, rather than the "Datasets" section. My reasoning is: the STL-Fashion/Home datasets are scene+product+bbox pairs. The subsequent postprocessing (cropping out the product from the scene image, etc) are specific to your training methodology - it's entirely possible that someone can train a CTL model on STL-Fashion/Home dataset without doing product-cropping, etc.
%WC: The generated CTL datasets are for both training and evaluation. Also the way to generated the dataset is one of the major contributions, putting it into the training section likely weakens its significance to me.
\subsection{Generating CTL Datasets}

%erickim: A figure that visualizes this postprocessing step would be invaluable to the reader. I see that you briefly visualize this step in Figure 2 (the one showing "Shop the Look" and "Complete the Look" data), but I think a figure dedicated to this crop operation will be helpful.
To generate CTL datasets based on STL data, and overcome the issue mentioned above, we propose to crop scene images to exclude their associated products. Given a scene image $I_s$ and a bounding box $B$ for a product, we consider four candidate regions (i.e.,~top, bottom, left, and right) that don't overlap with $B$, and select whichever has the greatest area as the cropped scene image.
Specifically, we perform the following procedure to crop scene images:
\begin{enumerate}[label=(\roman*),noitemsep,leftmargin=1.75\parindent,topsep=0pt]
%JULIAN: Description is somewhat vague. I get the idea but is there a reason to obscure the precise numbers? (Looks like it's described below, you might just mention that)
% \item Slightly expand the bounding box $B$ such that the product is likely to be fully covered.
\item In some cases, the bounding box doesn't fully cover the product. As we don't want the model to see even a small piece of the product (which might reveal e.g.~its color), we slightly expand the bounding box $B$ to ensure that the product is likely to be fully covered. Specifically, we expand all bounding boxes by
%a fixed margin (we use 
5\% of the image length.
\item Calculate areas of the candidate regions, and select the one whose area is largest. For fashion images,  we observe that almost all 
%the people 
subjects
are in a vertical pose, so we only consider the `top' and `bottom' regions 
%JULIAN: Added
(as the left/right regions often exclude the human subject);
for home images we consider all four candidates.
%JULIAN: Why the difference between fashion and home images?
\item Finally, as the cropped scene should be reasonably large so as to include the key context, we discard scene-product pairs for which the area of the cropped image is smaller than a threshold (we use 1/5 of the full area).
If the cropped image is large enough, the pair containing the cropped scene and the corresponding product is included in the CTL dataset.
\end{enumerate}

Following our
heuristic cropping strategy, we manually verify that in 
%JULIAN: Again, sounds vague. Concerning if it gives off the feeling that a large fraction of cropped images (e.g. 40%) *do* include the product
most cases the cropped image doesn't include the associated product, and the cropped image contains a meaningful and reasonable region for predicting 
%complementaries. 
complements.
%Also, 
In practice
we find 
%the
that
discarded 
%cases are largely due to 
instances are largely due to
dresses which often occupy a large area. 
%JULIAN: Why not? Dresses certainly have complements. I tried to rewrite this argument but please check.
%Moreover, for complement recommendation, it's not practical to recommend dresses. 
We find that for complement recommendation it is generally not practical to apply a cropping strategy for objects that occupy a large portion of the image;
therefore we opted simply to 
%remove 
discard
%all dress products. 
dresses from our dataset
(note that we can still recommend other fashion items based on scenes in which people wear dresses). Figure~\ref{fig:data} shows a comparison between STL and CTL data; CTL data statistics are listed in Table~\ref{tab:data}.

% train/test
% \vspace{-0.2cm}
\section{Method}\label{sec:method}

In the \emph{Complete the Look} task, we are given a dataset containing compatible pairs consisting of a scene image $I_s$ and a product image $I_p$ (as shown in Figure~\ref{fig:data}), and seek to learn scene-product style compatibility. To this end, we 
%specially 
design a model which measures the compatibility globally in addition to
%considering 
a more fine-grained approach that matches relevant regions of the scene image with the product image.

\subsection{Style Embeddings}

We adopt ResNet-50~\cite{DBLP:conf/cvpr/HeZRS16} to extract visual features from scene and product images. Based on the scene image $I_s$, we obtain a visual feature vector $\mathbf{v}_s\in\mathbb{R}^{d1}$ 
%JULIAN: Again, quite vague. If you really can't provide these details it's good to at least *sound* less vague. I tried to rewrite as such. It also mightn't be the worst idea to mention explicitly that these are proprietary details which you can't share.
%from last few hidden layers like \texttt{pool5}, 
from the final layers (e.g.~\texttt{pool5}),
and a feature map $\{\mathbf{m}_i\in\mathbb{R}^{d2}\}_{i=1}^{w\times h}$ from intermediate convolutional layers (e.g.~\texttt{conv4\_6}). Similarly, the visual feature for product image $I_p$ is denoted as $\mathbf{v}_p\in \mathbb{R}^{d1}$. Such ResNet feature vectors have shown strong performance and transferability~\cite{DBLP:conf/cvpr/HeZRS16,kornblith2018better}, and the feature maps have been shown to be able to capture key context from local regions~\cite{DBLP:conf/nips/RenHGS15,DBLP:conf/icml/XuBKCCSZB15}.

%erickim[2018-11-16] What is \hat{f}? how does it differ from {f}? Similarly, what is \hat{l} vs {l}?
Due to the limited size of our datasets, we freeze the weights of ResNet-50 (pretrained on Imagenet) and apply a two-layer feed forward network $g(\Theta;\cdot)$\footnote{The network architecture is \texttt{Linear-BN-Relu-Dropout-Linear\\-L2Norm}, and parameterized by $\Theta$.} to transform the visual features to a $d$-dimensional metric embedding (with unit length) in a unified style space. Specifically, we have:
\begin{equation}
\begin{split}
\mathbf{f}_s=g(\Theta_g;\mathbf{v}_s)&, \mathbf{f}_p=g(\Theta_g;\mathbf{v}_p),\\
\mathbf{f}_i=g(\Theta_l;\mathbf{m}_i)&, \mathbf{\hat{f}}_i=g(\Theta_{\hat{l}};\mathbf{m}_i),\\
\end{split}
\end{equation}
where $\mathbf{f}_s$ and $\mathbf{f}_p$ are the style embedding for the scene and the product respectively, and $\mathbf{f}_i, \mathbf{\hat{f}}_i$ are embeddings for the $i$-th region of the scene image. 
%The 
$\ell_2$ normalization is applied on embeddings to improve training stability, 
%which is 
an approach
commonly used in recent work on deep embedding learning~\cite{DBLP:conf/cvpr/SchroffKP15,DBLP:conf/iccv/ManmathaWSK17}.

\begin{figure}[t]
\centering
\includegraphics[width=\linewidth]{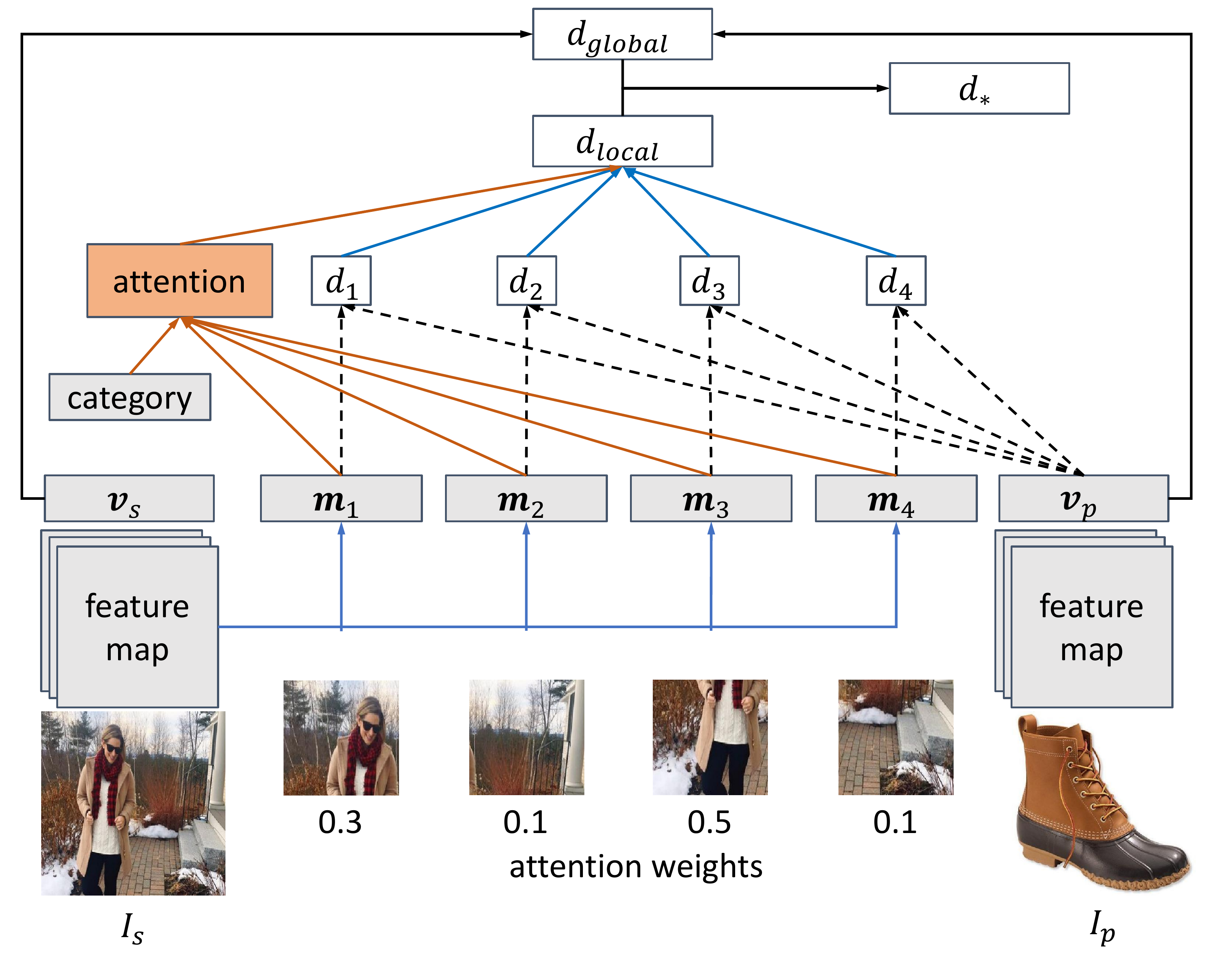}
\caption{An illustration of our hybrid compatibility measurement. We simplify the size of the attention map to 2$\times$2.}
\label{fig:method}
% \vspace{-0.2cm}
\end{figure}

\subsection{Measuring Compatibility}

We measure compatibility by considering both global and local compatibility in a unified style space.

%JULIAN: Switch to \paragraph{}? This renders kinda funny
\textbf{Global compatibility.} 
We seek to learn style embeddings from compatible scene and product images, where nearby embeddings indicate high compatibility. We use the (squared) $\ell_2$ distance between the scene embedding $\mathbf{f}_s$ and the product embedding $\mathbf{f}_p$ to measure their global compatibility:
\begin{equation}
d_{\text{global}} (s, p) = \|\mathbf{f}_s-\mathbf{f}_p\|^2,
\end{equation}
where $\|\cdot\|$ is the $\ell_2$ distance.

\textbf{Local Compatibility.}
As the scene image typically contains a large area including many objects, considering only global compatibility may overlook key details in the scene. Hence we match every region of the scene image with the product image to achieve a more fine-grained matching procedure. Moreover, not all regions are equally relevant, and relevance may vary when predicting complementarity from different categories. Thus, we first measure the compatibility between every scene patch and the product, and then adopt category-aware attention to assign weights over all regions:
\begin{equation}
\begin{split}
d_{\text{local}} (s, p) = \sum_{1\leq i\leq w\times h} a_{i}\|\mathbf{f}_{i}-\mathbf{f}_p\|^2,\\
\hat{a}_i= - \|\mathbf{\hat{f}}_{i} - \mathbf{\hat{e}}_c\|^2,\ \mathbf{a}= \text{softmax}(\mathbf{\hat{a}}),
\end{split}
\label{eq:local}
\end{equation}
where $c$ is the category of product $p$, and $\mathbf{\hat{e}}_c\in\mathbb{R}^d$ is an $\ell_2$-normalized embedding for category $c$. Here, we use the distance between $\mathbf{\hat{f}}_{i}$ and $\mathbf{\hat{e}}_c$ to measure the relevance of the $i$-th region of the scene image when predicting complements from category $c$. Note the `attentive distances' in eq.~\ref{eq:local} can be viewed as an extension of attention for metric embeddings, as if we replace the $\ell_2$ distance with an inner product we recover the conventional attention form $(\sum_i a_i\mathbf{f}_i)^T\mathbf{f}_p$.

%To summarize, 
Finally,
we measure compatibility by defining a hybrid distance that combines both global and local distances:
%JULIAN: Seems reasonable though would be nice to weight (e.g. softmax) over the two. It's unclear whether one of these two terms might dominate in practice when combining them with this simple weighting.
\begin{equation}
d_* (s, p) = \frac{1}{2}\left[d_{\text{global}} (s, p) + d_{\text{local}} (s, p)\right].
\end{equation}
Figure~\ref{fig:method} illustrates our scene-product compatibility measuring procedure.
Recall that all the 
%JULIAN: Is this actually a metric? Doesn't squaring distance mess up the triangle inequality? Not a big deal though.
embeddings we used are normalized to have unit length, and attention weights are also normalized (i.e.,~$\sum_i a_i=1$). 
%Hence, a nice property is that 
Note that
all the distances ($d_*$, $d_\text{glocal}$, and $d_\text{local}$)
%JULIAN: Worth mentioning? Not sure what the practical significance is. Could easily say more tersely (e.g.~"note that by construction 0 \leq d \leq 4")
range from 0~to~4.

\subsection{Objective}

Following~\cite{DBLP:conf/cvpr/SchroffKP15}, we adopt the hinge loss to learn style embeddings by considering triplets $\mathcal{T}$ of a scene $s$, a positive product $p^\text{+}$, and a negative product $p^\text{-}$:
\begin{equation}
\mathcal{L}= \sum_{(s,p^\text{+},p^\text{-})\in \mathcal{T}} \left[d_*(s, p^\text{+})-d_*(s, p^\text{-})+\alpha\right]_+,
\end{equation}
where $\alpha$ is the margin, which we set to 0.2 as in~\cite{DBLP:conf/cvpr/SchroffKP15}. Storing all possible triplets in $\mathcal{T}$ is intractable; we use mini-batch gradient descent to optimize the model, where training triplets for each batch are dynamically generated: we first randomly sample $(s,p^\text{+})$ from all compatible pairs, and then sample a negative product $p^\text{-}$ from the same category of $p^\text{+}$. We do not sample negatives from different categories, as during testing we rank products from the same category, 
%which is exactly the adopted sampling strategy seeks to optimize. 
%JULIAN: Wording is a bit borked, I don't completely follow
which is what the adopted sampling strategy seeks to optimize.

% For more detail refer to Section~\ref{sec:imp}, and to our supplementary material.

\begin{figure}[t]
\centering
\includegraphics[width=\linewidth]{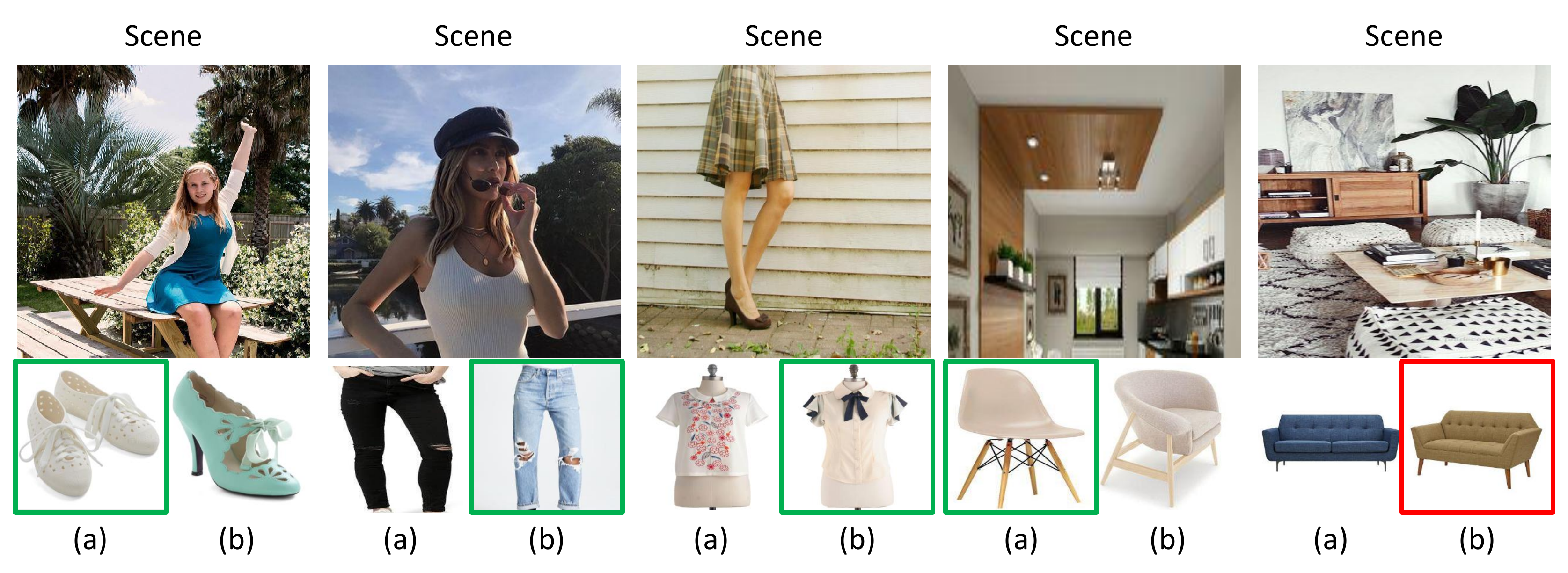}
\caption{A sample of the binary questions in our testing sets. Given a scene and two products, the model must predict which product is more compatible with the scene.
%JULIAN: In future use blue/yellow: (1) Red/green colorblindness is much more common; (2) Blue and yellow have different intensity and will be distinguishable if printed in black and white.
%The correct predictions are labeled with a green box, while red is for wrong answers.
Correct predictions are labeled in green, incorrect in red.
}
\label{fig:samples}
% \vspace{-0.2cm}
\end{figure}

\section{Experiments}\label{sec:exps}

% We first compare our approach against existing methods
% that are designed for predicting fashion compatibility between products. 
% We then study the behavior of the attention mechanism. Finally, we perform a human study and show 
% practical use-cases
% of our approaches. 

%JULIAN: I know this is criminally small but it was the easiest place to save space. I think if you up the font sizes a bit more you could maybe make this look nice

\subsection{Baselines}
\textbf{Popularity:} A simple baseline which recommends products based on their popularity (i.e., the number of associated $(\mathit{scene}, \mathit{product})$ pairs).

\textbf{Imagenet Features:} We directly use visual features from ResNet pretrained on Imagenet, which have shown strong performance in terms of retrieving visually similar images~\cite{DBLP:conf/cvpr/RazavianASC14,DBLP:conf/www/ZhaiKJFTDDD17}. The similarity is measured via the $\ell_2$ distance between embeddings.

\textbf{IBR}~\cite{DBLP:conf/sigir/McAuleyTSH15}: Image-based recommendation~(IBR) measures product compatibility via a learned Mahalanobis distance between visual embeddings. Essentially IBR learns a linear transformation to convert visual features into a style space.

\textbf{Siamese Nets:} Veit et al.~\cite{DBLP:conf/iccv/VeitKBMBB15} adopt Siamese CNNs~\cite{DBLP:conf/cvpr/ChopraHL05} to learn style embeddings from product images, and measure their compatibility using an $\ell_2$ distance. As suggested in \cite{DBLP:conf/iccv/VeitKBMBB15}, we fine-tune the network based on a pretrained model.

\textbf{BPR-DAE}~\cite{DBLP:conf/mm/SongFLLNM17}: This method uses autoencoders to extract representations from clothing images and textual descriptions, and incorporates them into the BPR recommendation framework~\cite{DBLP:conf/uai/RendleFGS09}. Due to the absence of textual information in our datasets, we only use its visual module.

Since the baselines above are designed for measuring product compatibility, we adapt the baselines to our problem by treating all images as product images and apply the same sampling strategy as used in our method.
%JULIAN: Don't write (eq.xx)! Better to write (eq.\ref{xx}) so that it at least throws an error until you fill it in.

%JULIAN: Baselines mostly good, hard to say how fair some of these are, the above limitation makes it look like the baselines will be a bit broken due to an image heterogeneity issue that they weren't designed for. Would also potentially be nice to have variants of your own approach to understand the influence of different components or alternative choices.

\subsection{Implementation Details}
\label{sec:imp}

For a fair comparison, we implemented all methods using ResNet-50\footnote{We use the implementation from TensorFlow-Slim. The architecture is slightly different from the original paper (e.g. different strides).} (pretrained on Imagenet),
as the underlying network, 
where the layer \texttt{pool5}~(2048d)
%JULIAN: Is this fair? Doesn't your method use additional layers as well as pool5?
is used for the visual vectors and the layer \texttt{block3}~ (7$\times$7$\times$1024) is used as the feature map. We use an embedding size of
128, and we did not observe any performance gain with larger $d$ (e.g.~$d=512$). All models are trained using \emph{Adam}~\cite{DBLP:journals/corr/KingmaB14} with a batch size of 16. As suggested in~\cite{DBLP:conf/cvpr/SchroffKP15}, visual embeddings are normalized to have a unit length for metric embedding based methods, and the margin $\alpha$ is set to $0.2$. For all methods, horizontal mirroring and random $224\times224$ crops from $256\times256$ images are used for data augmentation, and a single center crop is used for testing. We randomly split the scenes (and the associated pairs) into training (80\%), validation (10\%) and test (10\%) sets. We train all methods for 100 epochs, examine the performance on the validation
%and test 
set every 10 epochs, and report the test performance for the model which achieves the best validation performance.

\subsection{Recommendation Performance}

As shown in Figure~\ref{fig:samples}, given a scene $s$, a category $c$, a positive product $p^\text{+}$, and a negative product $p^\text{-}$ (randomly sampled from $c$), the model needs to decide which product is more compatible with the scene image. 
% Following~\cite{DBLP:conf/mm/HanWJD17,DBLP:conf/mm/SongFLLNM17}, we assume the randomly sampled product $p^\text{-}$ is generally less compatible than $p^\text{+}$. 
The accuracy of these binary choice problems is used as a metric for performance evaluation.
%JULIAN: Which baselines are optimizing binary accuracy metrics, and which are optimizing contrastive losses?
Note the 
%JULIAN: Seems a bit risky to mix the word "accuracy" with something that really corresponds to the AUC. Could rewrite like "Note that the fraction of times that we correctly resolve these binary choice problems is equivalent to the AUC" or something.
accuracy here is equivalent to the AUC which measures the overall ranking performance.

\begin{figure}[t]
\centering
\begin{subfigure}[b]{0.33\linewidth}
\centering
\includegraphics[width=\linewidth]{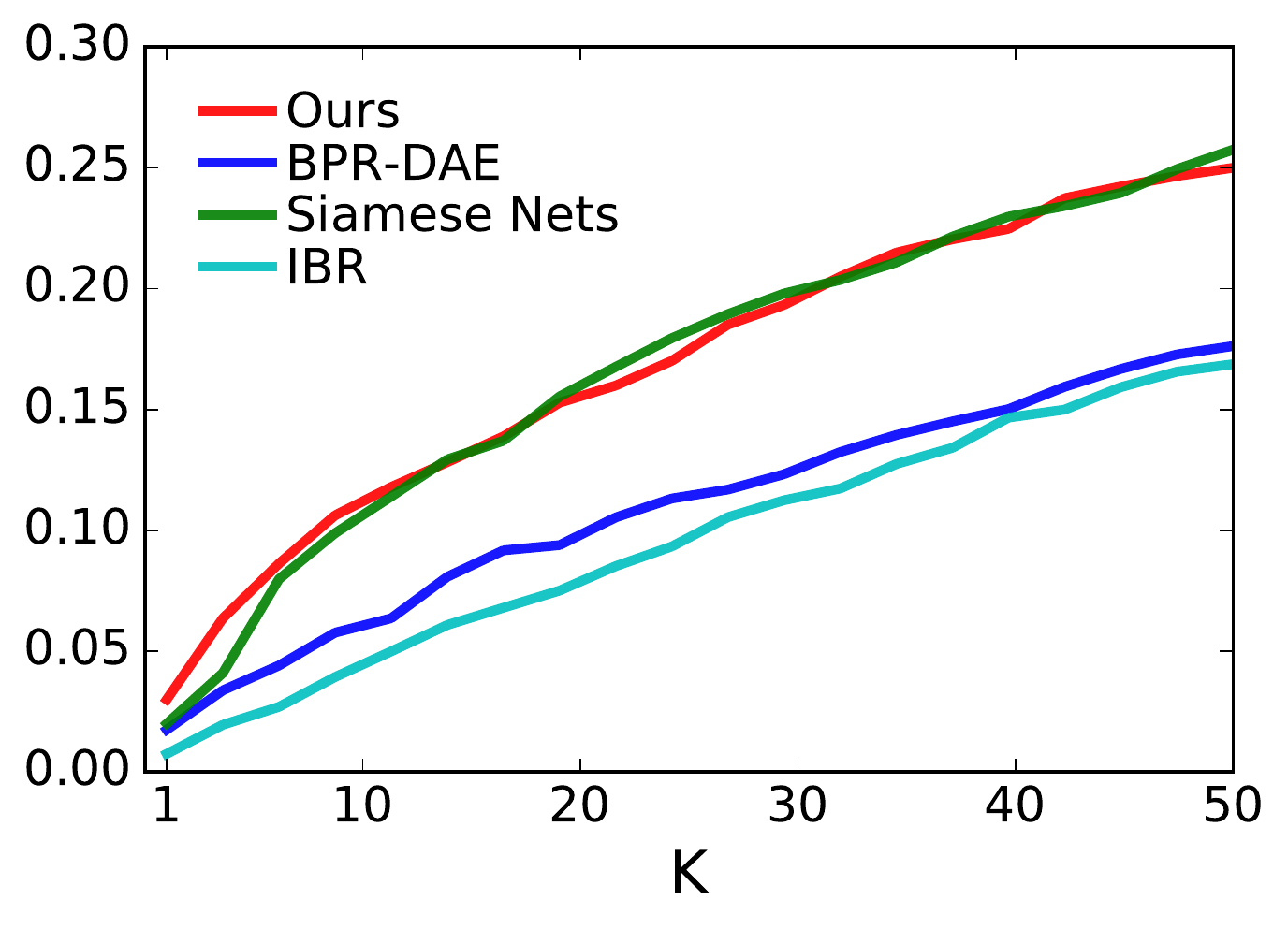}
% \vspace{-0.2cm}
\subcaption{\textbf{Fashion-1}}
\end{subfigure}%
\begin{subfigure}[b]{0.33\linewidth}
\centering
\includegraphics[width=\linewidth]{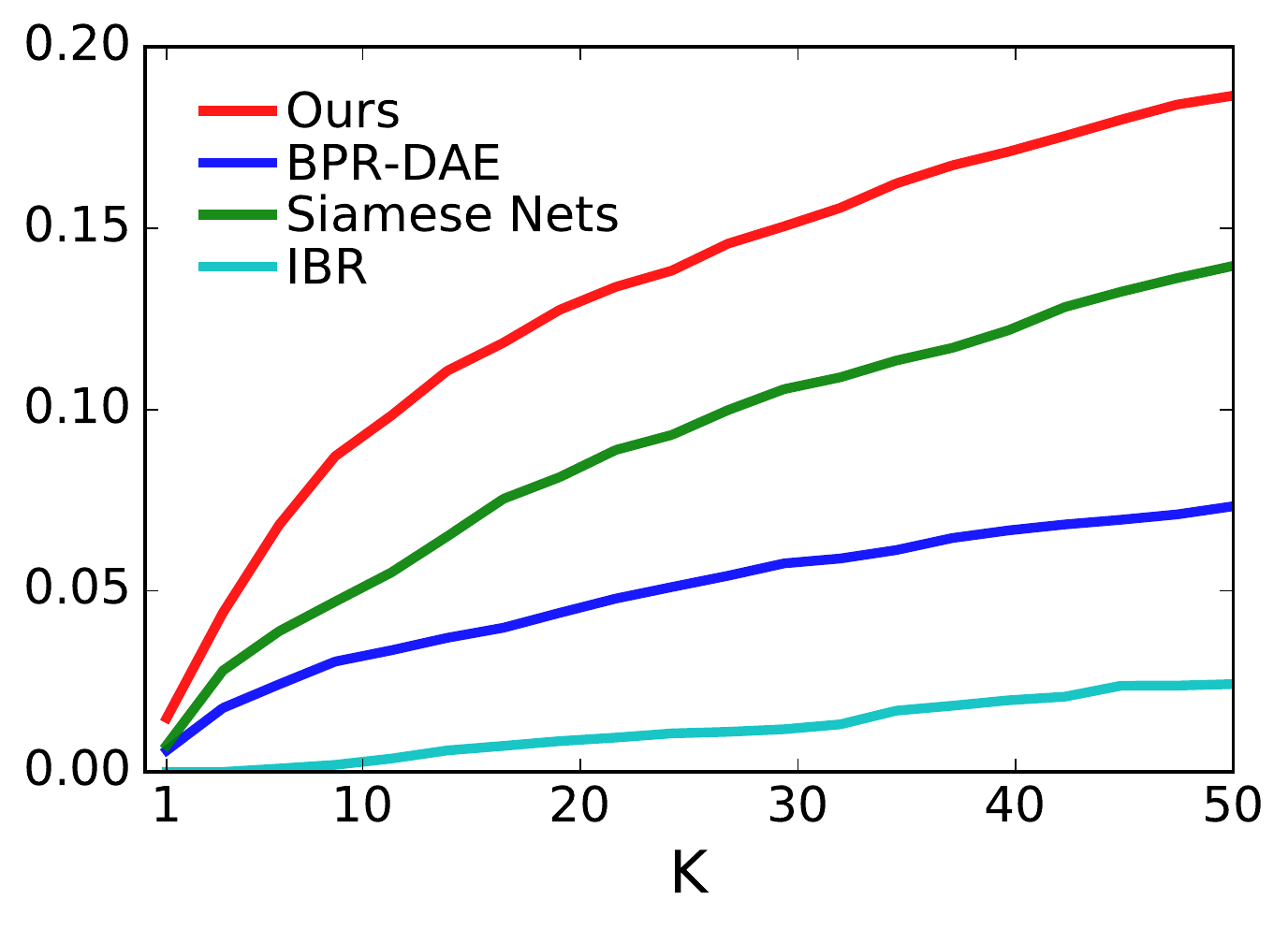}
% \vspace{-0.2cm}
\subcaption{\textbf{Fashion-2}}
\end{subfigure}
\begin{subfigure}[b]{0.33\linewidth}
\centering
\includegraphics[width=\linewidth]{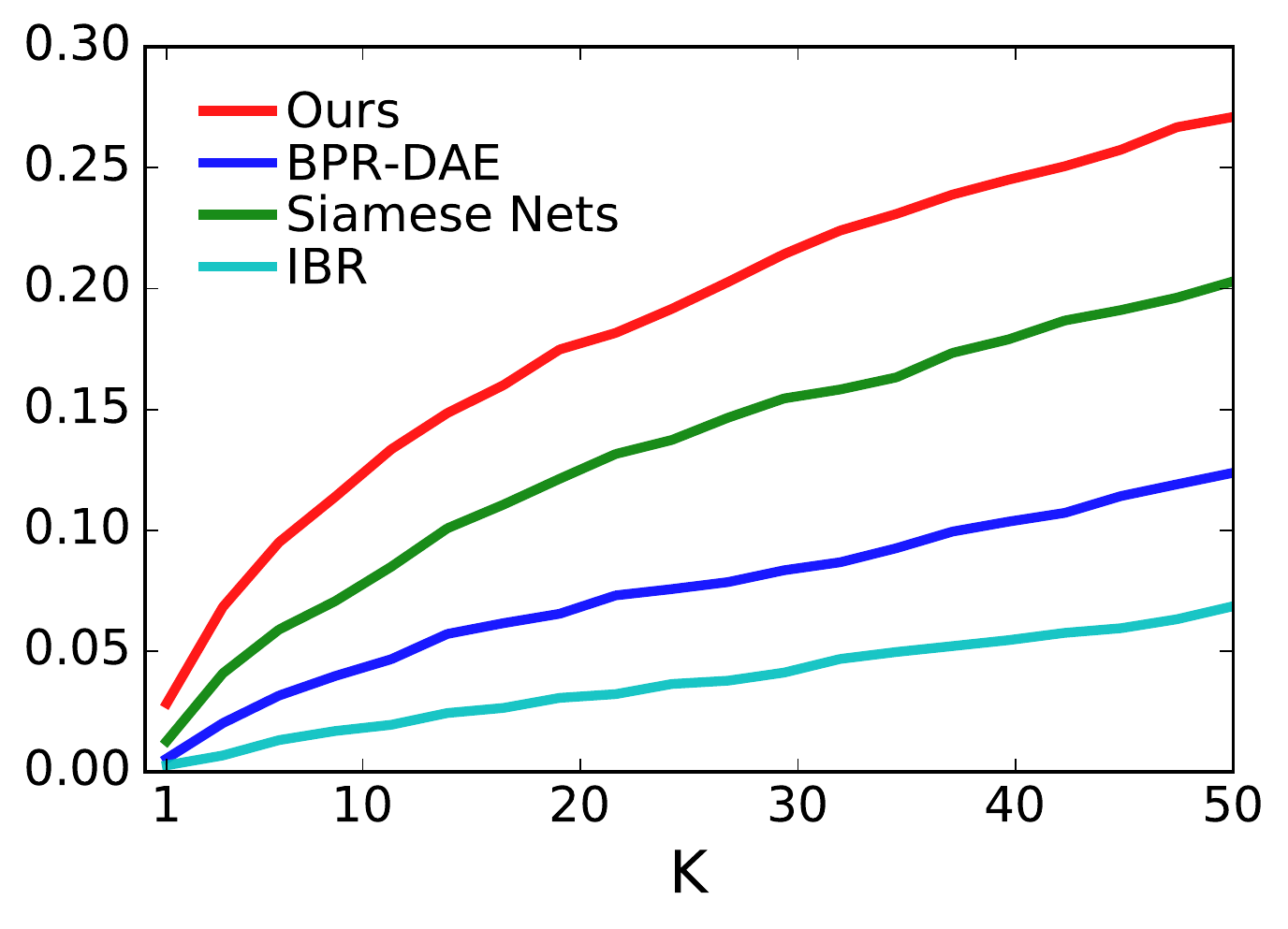}
% \vspace{-0.2cm}
\subcaption{\textbf{Home}}
\end{subfigure}
% \includegraphics[width=.25\linewidth]{figs/TopK_ESTS}
% \includegraphics[width=.25\linewidth]{figs/TopK_Fashion}
% \includegraphics[width=.25\linewidth]{figs/TopK_Home}
%JULIAN: Somewhat unreadable figure due to font sizes
\caption{Top-K Accuracy on all datasets (i.e.,~how often the top-K retrieved items contain the ground-truth product).}
\label{fig:topk}
% \vspace{-0.2cm}
\end{figure}

\begin{table}[h]
\small
\centering
\begin{tabularx}{.95\linewidth}{l *3{>{\centering\arraybackslash}X}@{}}
% \toprule
\textbf{Method}														&	\textbf{Fashion-1}	&	\textbf{Fashion-2}	&	\textbf{Home}\\	\hline
Random 													&	50.0	&	50.0	&	50.0	 \\
Popularity 												&	52.1	&	57.5	&	55.6	 \\
Imagenet Feature										&	49.4	& 	51.6	&	48.1 \\                         
\hline
\multicolumn{4}{l}{\emph{Train w/ full images}}\\
IBR~\cite{DBLP:conf/sigir/McAuleyTSH15} 				&	56.5	& 58.5	&	57.0 \\     
Siamese Nets~\cite{DBLP:conf/iccv/VeitKBMBB15}			&	63.0	& 67.1	&	72.4 \\                      
BPR-DAE~\cite{DBLP:conf/mm/SongFLLNM17}					&	59.3	& 61.1	&	64.2 \\ 
Ours													&	63.1	& 70.0	&	75.0 \\ 
\hline
\multicolumn{4}{l}{\emph{Train w/ cropped images}}\\
IBR~\cite{DBLP:conf/sigir/McAuleyTSH15} 				&	54.5	& 55.9	&	58.0 \\     
% IBR~\cite{DBLP:conf/sigir/McAuleyTSH15}, d=512 			&	00.0	& 00.0	&	00.0 \\     
Siamese Nets~\cite{DBLP:conf/iccv/VeitKBMBB15}			&	64.0	& 69.0	&	73.1 \\         
% Siamese Net~\cite{DBLP:conf/iccv/VeitKBMBB15}, d=512			&	00.0	& 00.0	&	00.1 \\
BPR-DAE~\cite{DBLP:conf/mm/SongFLLNM17}					&	59.6	& 61.1	&	65.8 \\
% BPR-DAE~\cite{DBLP:conf/mm/SongFLLNM17}, d=512						&	00.0	& 00.0	&	00.0 \\
Ours													&	\textbf{68.5}	&	\textbf{75.3}	&	\textbf{79.6}	 \\
% \bottomrule
\end{tabularx}
\caption{Accuracy of  binary comparisons on all datasets.}
\label{tab:auc}
% \vspace{-0.2cm}
\end{table}

Table~\ref{tab:auc} lists the 
%JULIAN: Again, not sure whether this is really "accuracy" or AUC, or even if it's accuracy for some baselines but AUC for others.
accuracy of all methods. First, we note that the first group of methods (learning-free) perform poorly.
%on this task. 
Imagenet features perform similarly to random guessing, which indicates that visual compatibility is 
%JULIAN: I don't really believe this, or at least I might tone it down -- I suspect it's mostly because you're comparing a scene image to a product image, but the non-learning methods can't learn to ignore the background. It's entirely possible that naive similarity *would* work well if not for this image heterogeneity issue. I might simply say that the non-learning methods can't work when faced with a scene image and a product image, and not focus too much on the similarity angle. There's a small but non-zero risk of upsetting authors of competing methods by showing that their techniques don't work, without re-iterating that the reason they don't work is mostly because the data is incompatible with their assumptions (rather than due to a fundamental flaw in their methods)
different from visual similarity, 
and thus it is necessary to learn the notion of compatibility from data. Even the na\"ive popularity baseline achieves better (though still poor) performance. Second, we found training with
%JULIAN: I would explicitly mention that you're going to perform this comparison when you discuss the baselines above.
cropped images is effective as it can generally boost the performance compared with using full images. Compared to other baselines, our method has the most significant performance drop with full images, presumably because our method is the only 
%local-aware 
one which is aware of local appearance,
%method which is 
which makes it easier to erroneously match
%easier to match 
scene patches with the product rather than leaning compatibility (as discussed in Section~\ref{sec:issue}). The performance gap between Fashion-1 and Fashion-2 possibly relates to their sizes. Finally, our method achieves the best performance on all datasets for both the fashion and home domains.

In addition to an overall ranking measurement, the Top-K accuracy (the fraction of times that the first K recommended items contain the positive item)~\cite{ak2018learning} might be closer to a practical scenario. Figure~\ref{fig:topk} shows Top-K accuracy curves for all datasets. We see that our method significantly outperforms baselines on the last two datasets, and slightly outperforms the strongest baseline on the first dataset.
%JULIAN: Not sure I follow. Wouldn't this only change the absolute Top-K performance, and not the *relative* performance compared to baselines?
% Presumably this is related to the size of datasets as the 
%JULIAN: Might be better to name them rather than calling them "first dataset" etc.
% first dataset 
% has a very limited number of products for retrieval.
%JULIAN: This is pretty important and too easy for the reviewers to miss. Would be nice to say *what* model variants are compared in the supplementary material. I also wonder if there might be one or two worth promoting to the main text, especially if it's just a single paragraph and a line in the table.
Performance analysis on additional model variants is included in our supplementary material.

\subsection{Analysis of Attended Regions}

We visualize the attended areas to intuitively reveal what parts of a scene image are important for predicting complementarity, and quantitatively evaluate whether the attention focuses on meaningful regions.

\begin{figure}[t]
\includegraphics[width=\linewidth]{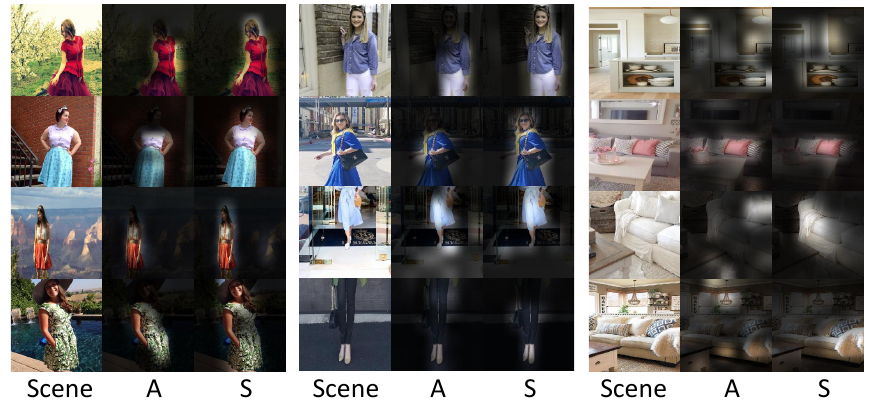}
\caption{Visualization of attention maps (`A') from our method, and saliency maps (`S') from DeepSaliency~\cite{DBLP:journals/tip/LiZWYWZLW16}.}
\label{fig:att}
% \vspace{-0.2cm}
\end{figure}

Figure~\ref{fig:att} shows test scene images (after cropping), the corresponding attention map from our model, and the saliency map generated by DeepSaliency\footnote{\url{http://www.zhaoliming.net/research/deepsaliency}}~\cite{DBLP:journals/tip/LiZWYWZLW16}. DeepSaliency
%JULIAN: Text was borked, not sure if something is missing here?
%a and 
is trained to detect salient objects while our attention mechanism discovers relevant areas by learning the compatibility between scene and product images. For the first two fashion datasets, the two approaches both successfully identify the subject of the image (i.e.,~the person) from various backgrounds. Interestingly, our attention mechanism tends to ignore human faces and more focus only on clothing, which means our model 
%thinks the 
discovers that the subject's clothing is more relevant than the appearance of the subject themselves
%clothing you wear are more relevant 
when 
%making complementary recommendation. 
recommending complements.\footnote{Note that attention is only used when measuring local compatibility, the model can still leverage the context provided in the unattended regions via the global compatibility.}
In contrast, the scenes in the interior design domain are much more complex
%(e.g.~containing more objects). 
(critically, they contain many objects rather than a single subject).
Although some meaningful objects (e.g.~pillows, lamps, etc.) are discovered in some cases, it appears to be harder for 
%both methods 
either attention or saliency
to detect key objects.

% \begin{table}[t]
% \small
% \centering
% \begin{tabularx}{.95\linewidth}{l *3{>{\centering\arraybackslash}X}@{}}
% \textbf{Method}													& \textbf{Fashion-1}					&	\textbf{Fashion-2}	&	\textbf{Home}\\	\hline
% \multicolumn{4}{l}{\emph{Top-1 region}}\\
% Random																&	13.2 						&	12.3						& 16.4	 \\
% \multirow{2}{*}{Ours}												&	22.4						&	24.4						& 18.9	\\
% 																	&	(+70\%)						&	 (+98\%)					&  (+15\%)\\
% \multirow{2}{*}{DeepSaliency~\cite{DBLP:journals/tip/LiZWYWZLW16}}	&	24.8						&	25.0						& 17.8  	 \\
% 																	&	(+88\%)						&	 (+103\%)					& (+8.5\%) 	 \\
% \hline
% \multicolumn{4}{l}{\emph{Top-3 regions}}\\
% Random																&	32.1 						&	29.9						&	37.0 \\
% \multirow{2}{*}{Ours}												&	43.3 						&	45.0 						& 	38.3 \\
% 																	&	(+35\%)						&	(+51\%) 					& (+3.5\%)\\
% \multirow{2}{*}{DeepSaliency~\cite{DBLP:journals/tip/LiZWYWZLW16}}	&	49.2						&	47.8 						& 36.8\\
% 																	&	(+53\%)						&	(+60\%)						& (-0.5\%)
% \end{tabularx}
% \caption{Does the attention focus on meaningful regions?}
% \label{tab:att}
% \end{table}

\begin{table}[t]
\small
\centering
\begin{tabularx}{.95\linewidth}{l *3{>{\centering\arraybackslash}X}@{}}
\textbf{Method}													& \textbf{Fashion-1}					&	\textbf{Fashion-2}	&	\textbf{Home}\\	\hline
\multicolumn{4}{l}{\emph{Top-1 region}}\\
Random																&	13.2 						&	12.3						& 16.4	 \\
Attention (Ours)												&	22.4						&	24.4						& 18.9	\\
																	
DeepSaliency~\cite{DBLP:journals/tip/LiZWYWZLW16}	&	24.8						&	25.0						& 17.8  	 \\
																	
\hline
\multicolumn{4}{l}{\emph{Top-3 regions}}\\
Random																&	32.1 						&	29.9						&	37.0 \\
Attention (Ours)												&	43.3 						&	45.0 						& 	38.3 \\
																	
DeepSaliency~\cite{DBLP:journals/tip/LiZWYWZLW16}	&	49.2						&	47.8 						& 36.8\\
\end{tabularx}
% \caption{How often does the attention and saliency map focus on meaningful regions?}
\caption{Fraction of successful hits on meaningful regions.}
\label{tab:att}
% \vspace{-0.2cm}
\end{table}

In addition to qualitative examples, we also quantitatively measure whether attention focuses on meaningful areas of scene images. Here we 
%JULIAN: Didn't quite follow, tried to rewrite
%think the 
assume that
areas 
corresponding to
%of 
labeled products are relevant. Specifically, we divide the image into $7\times7$ regions, and a region is considered `relevant' if it 
%JULIAN: Vague. Does this mean more than 50%? Without knowing quite what's being measured here it's a little difficult to trust the evaluation.
%largely 
significantly
overlaps (i.e. larger than half the area of a region) with any product's bounding box. We then calculate the attention map ($7\times 7$) for all test scene images and rank the 49 regions according to their scores. If the top-1 region 
%JULIAN: Seems redundant, can you not just say you evaluate the relevance measure in terms of its Top-3 performance?
(or one of the top-3 regions) is a `relevant region', then we deem the attention map as a successful hit. 

Table~\ref{tab:att} shows the fraction of successful hits of our attention map, random region ranking, and the saliency map from DeepSaliency~\cite{DBLP:journals/tip/LiZWYWZLW16}. 
%For the first two 
On the
fashion datasets, our method's performance is close to that of DeepSaliency, and both methods are significantly better than random. This shows that our attention mechanism can discover and focus on key objects (without knowing what area is relevant during training) guided by the supervision of complementarity.
% and tends to focus on them rather than less relevant areas like the image background. 
% WC: verbose
This is similar to a recent study which shows that spatial attention seems to be good at extracting key areas for clothing category prediction~\cite{wang2018attentive}.
%JULIAN: Didn't follow what you meant by saying you're the first? I don't see the difference.
%, though we first quantitatively study the attention behavior in the fashion domain.
% and fashion retrieval~\cite{DBLP:conf/kdd/ZhangPZZZRJ18}. 
However, for the `home' domain, both our method and DeepSaliency perform only slightly better than (or similar to) random. This indicates that scene images in the home domain are more complex than fashion images, as shown in Figure~\ref{fig:att}. This may also imply that a more sophisticated method
%a better way~
(e.g.~object detection) might be needed to extract local patterns for the home domain.

%erickim[2018-11-16] We probably won't have time to do this for the paper, but a really cool figure would be to visualize qualitative results for the same person+outfit, but different contexts (backgrounds). In this way, we could demonstrate how the model is recommending bags/hats/shoes/etc that complement the scene (context). One way to do this is to copy+paste the segmented person onto different scenes (beach, nature, indoor, etc).
\begin{figure*}
\includegraphics[width=.99\linewidth]{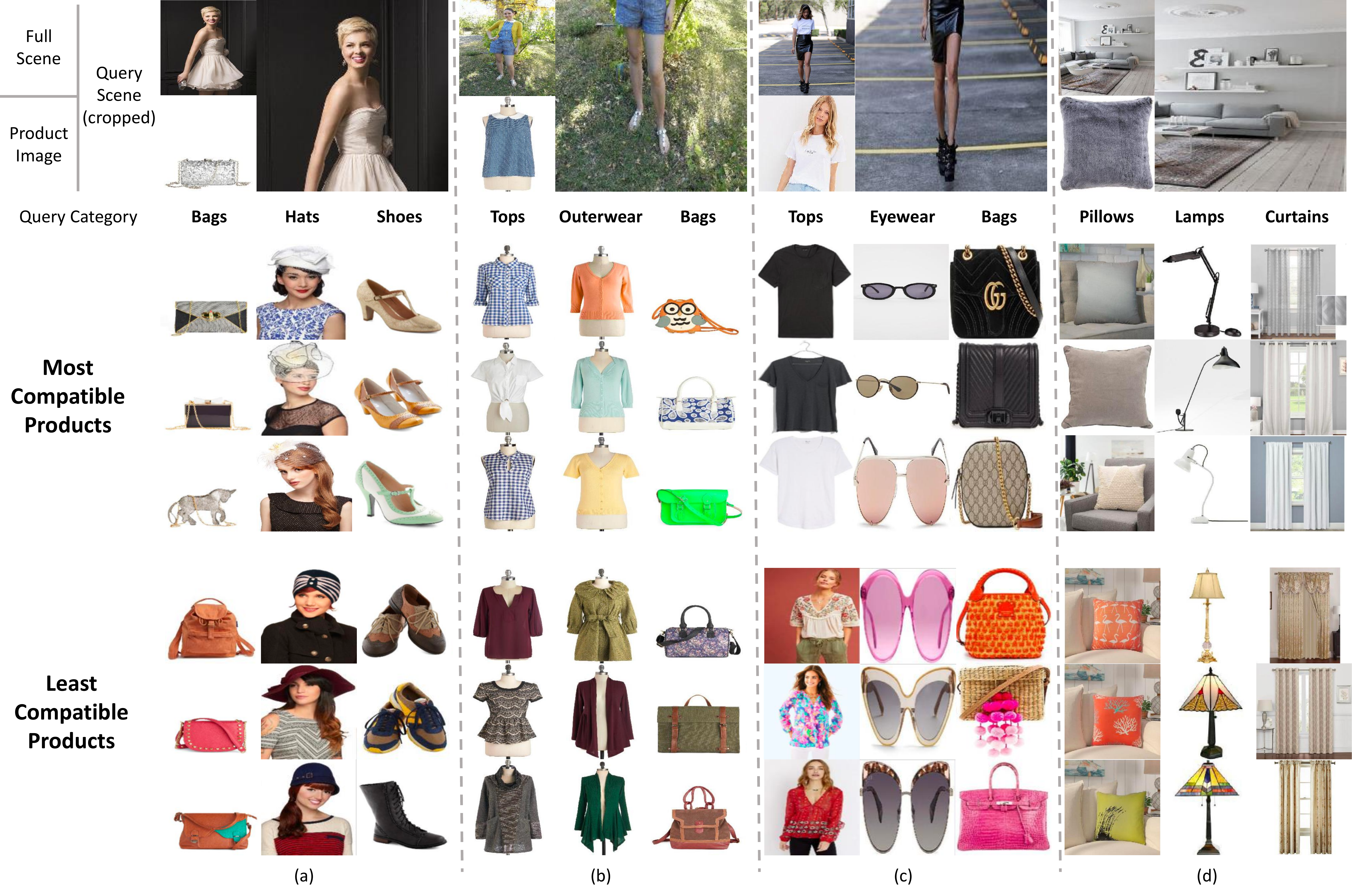}
\caption{Qualitative results of the top-3 most and least compatible products generated by our model. Note the full scenes and (ground truth) product images are only for demonstration and are not the input to our system.}
\label{fig:examples}
% \vspace{-0.2cm}
\end{figure*}

\begin{table}[t]
\small
\centering
\begin{tabularx}{.95\linewidth}{l *3{>{\centering\arraybackslash}X}@{}}
% \toprule
\textbf{Method}												& \textbf{Dataset}								&	\textbf{Overlap}					&	\textbf{Human}\\ 
                                                \hline
Popularity											&	56.5 		&	60.7	&	56.0 	 \\        
IBR~\cite{DBLP:conf/sigir/McAuleyTSH15} 			&	56.3		&	56.2	&	52.9   \\
Siamese Nets~\cite{DBLP:conf/iccv/VeitKBMBB15}		&	72.1 		&	72.6	&	62.1 \\
BPR-DAE~\cite{DBLP:conf/mm/SongFLLNM17}				&	62.5 		&	63.8	&	58.3  \\ 
Ours												&	\bftabnum 75.8 			&	\bftabnum 77.3		&	\bftabnum 65.0       	 \\
\hline
Human&	75.0	& 100	&	100 \\
% \bottomrule
\end{tabularx}
\caption{Accuracy on sampled data using dataset labels, human labels, and overlap labels as ground truth (respectively).}
\label{tab:human}
% \vspace{-0.2cm}
\end{table}

\subsection{Human Performance}

To assess how well the learned models accord with human fashion sense, we conduct a human subject evaluation, in which four fashion experts are asked to respond to binary choice questions (as shown in Figure~\ref{fig:samples}).
Specifically, each fashion expert is required to label 20 questions (randomly sampled from the test set) for each dataset, and performance is then evaluated based on the 240 labeled questions. 
%Table~\ref{tab:human} lists the accuracy using dataset labels, human labels, and overlapping labels (i.e., cases where the human and dataset have the same label) as ground truth labels.
%Table~\ref{tab:human} lists the accuracy of different evaluation settings: using dataset labels, human labels, and overlapping labels (i.e., cases where the human and dataset have the same label) as ground truth labels.
%JULIAN: Not really sure what labels are, maybe that will become clear when the figure is added, but would still be good for this paragraph to be self-contained
%With dataset labels, human accuracy is 75\%, which
%reflects the overall difficulty (or ambiguity) of the task.
%means the task is hard as it's far below 100\%. 
%We analyze the disaccord cases in supplementary material. 
%We can observe the positive correlation between the performance with dataset labels and human labels: better performance on the dataset typically implies better consistency with human fashion sense. 
%With dataset labels, our method is slightly better than human performance,
%which in turn
%is better than other baselines. Our method also achieves the best performance with overlapping labels and human labels. By conducting significance tests, our method is significantly better\footnote{$p<0.05$, by Fisher's exact test~\cite{agresti1992survey}.} than the baselines except Siamese Nets~\cite{DBLP:conf/iccv/VeitKBMBB15}, presumably due to the limited sample size.
An important observation is that our model achieves `human-level' performance: the second column of Table~\ref{tab:human} (``Dataset'') shows that fashion experts achieve 75.0\% accuracy, while our model achieves 75.8\% accuracy.

% Considering that fashion experts sometimes disagree with the ground-truth, a natural question to ask is: how does our model perform, relative to the fashion experts' judgments? 
% In this scenario, we use the fashion experts' judgments as the ground-truth labels, and our model outperforms other methods (fourth column of Table~\ref{tab:human}, ``Human''). 
Considering that fashion experts sometimes disagree with the ground-truth, we use the fashion experts' judgments as the ground-truth labels to evaluate the consistency with human fashion sense, and our model outperforms other methods (fourth column of Table~\ref{tab:human}, ``Human''). 
We also observe that better performance on the dataset typically implies a better consistency with human fashion sense.
%WC: shorten a bit and add the consistency between dataset and human to show the usefulness of the generated datasets. 

A final question is related to the subjectivity of the task: how well does our model do on cases where there is a clear answer?
% Finally, we investigate the subjectivity of the task by generating a dataset consisting of only unambiguous questions: select the test data where both fashion experts and the dataset label agree.
To answer this question, we used the following heuristic to generate a dataset consisting of only unambiguous questions: select the test data where both fashion experts and the dataset label agree.
On this data subset, our model is again the top-performer, which shows that our model indeed produces better fashion recommendations than other approaches, even when controlling for question ambiguity (third column of Table~\ref{tab:human}, ``Overlap'').

% \begin{table}[h]
% \small
% \centering
% % \centering\sisetup{table-format=1.4, table-number-alignment=center}
% \begin{tabular}{l *3 c}
% % \toprule
% \textbf{Method}												& \textbf{Dataset}								&	\textbf{Overlap}					&	\textbf{Human}\\ 
% % \# of samples									&	240								&   180						&	240\\                                             
%                                                 \hline
% POP 												&	56.5\rlap{$^{**}$} 		&	60.7\rlap{$^{**}$}	&	56.0\rlap{$^{**}$} 	 \\        
% IBR~\cite{DBLP:conf/sigir/McAuleyTSH15} 			&	56.3\rlap{$^{**}$}		&	56.2\rlap{$^{**}$}	&	52.9\rlap{$^{**}$}   \\     
% Siamese Net~\cite{DBLP:conf/iccv/VeitKBMBB15}		&	72.1 					&	72.6				&	62.1 \\         
% BPR-DAE~\cite{DBLP:conf/mm/SongFLLNM17}				&	62.5\rlap{$^{**}$} 		&	63.8\rlap{$^{**}$}	&	58.3\rlap{$^{*}$}  \\ 
% Ours												&	\bftabnum 75.8 			&	\bftabnum 77.3		&	\bftabnum 65.0       	 \\
% \hline
% Human&	75.0	& 100	&	100 \\
% % \bottomrule
% \end{tabular}
% \caption{Human Evaluation. *$p<0.1$, **$p<0.05$}
% \end{table}

\subsection{Qualitative Results}

%JULIAN: Changed this paragraph a lot. The results were convincing
Figure~\ref{fig:examples} shows four examples
(from the test set)
that depict the original scene, 
%removed 
the cropped
product, query scene, query category, and the top-3 most and least compatible products 
selected
by our algorithm. By comparing the removed product and the retrieved products from the first category (i.e.,~the same as the removed one),
%JULIAN: Added, as the statement, while true, is purely subjective
it appears that
the most compatible items 
are closer in style to
%more similar 
the ground-truth product compared with the least compatible items. 
%JULIAN: "intuitively" seems the wrong word
%Intuitively speaking, 
Qualitatively speaking,
the generated compatible products are 
%generally 
more compatible with the scenes. In column (a), the recommended white and transparent hats are (in the authors' opinion) more compatible
%JULIAN: Some of the statements are too qualitative, need to distinguish between opinion and fact
%with the lady than the hats with 
than
dark colors; in column (b), 
%we can see a women wearing a 
the
yellow 
%outwear 
outerwear
from the full scene
is close in color and style with the recommendations.
%, and the third recommended outwear looks pretty similar. 
We also observe that the learned compatibility is not merely based on 
%some 
simple factors like color;
for example, in column (d) the recommended lamps have different colors but similar style (modern, minimalist), and are quite different in style from the incompatible items. Thus the model appears to have learned a complex notion of style.
%For example, the model recommends both black and white lamps in the case (d)
%. However, the recommended lamps seem to have similar styles (e.g. simple, modern), while the incompatible lamps seem to belong another kind of style. 
%This shows the model learned a complex notion of style.

%------------------------------------------------------------------------
\section{Conclusion}

%erickim[2018-11-16] The phrase "automatically determine absent categories, instead of manually indicating them" seems out-of-place and unexpected, with little context (ie I'm not sure what it means). Can we remove it?
%JULIAN: I removed -- was one line too long anyway
In this paper, we proposed a novel task, \emph{Complete the Look}, for recommending complementary products given a real-world scene. Complete the Look (or CTL) can be straightforwardly applied on e-commerce websites to give users fashion advice, simply by providing scene images as input. We designed a cropping-based approach to construct CTL datasets from STL (Shop the Look) data. We estimate scene-product compatibility globally and locally via a unified style space. We performed extensive experiments on recommendation performance to verify the effectiveness of our method. We further studied the behavior of our attention mechanism across different domains, and conducted human evaluation to understand the ambiguity and difficulty of the task. Qualitatively, our CTL method generates compatible recommendations that seem to capture a complex notion of `style.' In the future, we plan to incorporate object detection techniques to extract key objects for more fine-grained compatibility matching.
%, and automatically determine absent categories, instead of manually indicating them.

\section*{Acknowledgement} The authors would like to thank Ruining He, Zhengqin Li, Larkin Brown, Zhefei Yu, Kaifeng Chen, Jen Chan, Seth Park, Aimee Rancer, Andrew Zhai, Bo Zhao, Ruimin Zhu, Cindy Zhang, Jean Yang, Mengchao Zhong, Michael Feng, Dmitry Kislyuk, and Chen Chen for their help in this work.
{\small
\bibliographystyle{ieee}
\bibliography{egbib}
}

\end{document}

%%%%%%%%% TITLE
\title{Complete the Look: Scene-based Complementary Product Recommendation\\(Supplementary Material)}

\author{Wang-Cheng Kang\textsuperscript{\dag}\thanks{Work done while intern at Pinterest.} , Eric Kim\textsuperscript{\ddag}, Jure Leskovec\textsuperscript{\ddag\S}, Charles Rosenberg\textsuperscript{\ddag}, Julian McAuley\textsuperscript{\dag}\\
\textsuperscript{\ddag}Pinterest, \textsuperscript{\S}Stanford University, \textsuperscript{\dag}UC San Diego\\
{\tt\small \{wckang,jmcauley\}@ucsd.edu, \{erickim,jure,crosenberg\}@pinterest.com}
}


\maketitle
\thispagestyle{empty}

%%%%%%%%% BODY TEXT
\section{Performance Analysis}
\subsection{Ablation Study}

We perform an ablation study to analyze the effect of the global and local compatibility measuring components. Table~\ref{tab:abla} shows the accuracy of our method and three variants on all datasets. `L' and `G' represents the variants with only the local component and the global component (respectively). We also exam the performance of the variant `G+L\textsuperscript{0}' which assigns equal weights on each region (instead of using attention weights). `G+L' is the default model which uses both components and attention weights. We can see the hybrid model `G+L' is better than using the two components individually. Also, the attention is helpful as it can boost performance compared with `G+L\textsuperscript{0}'.
\begin{table}[ht]
\small
\centering
\begin{tabularx}{\linewidth}{l *3{>{\centering\arraybackslash}X}@{}}
% \toprule
\textbf{Method}														&	\textbf{Fashion-1}	&	\textbf{Fashion-2}	&	\textbf{Home}\\	\hline
L												&	67.6	&	73.8	& 77.1 	 \\
G												&	66.9	&	74.2	& 77.8	\\
G+L\textsuperscript{0}							&	66.9	&	74.1		& 78.6	 \\
\hline
G+L (Default)									&	\textbf{68.5}	&	\textbf{75.3}	&	\textbf{79.6}	 \\
\end{tabularx}
\caption{Ablation Study}
\label{tab:abla}
\end{table}

\subsection{The Effect of Local Features}

As our method utilizes intermediate feature maps as local features, we exam the effect of using features from different layers and networks. In addition to different blocks of ResNet-50~\cite{DBLP:conf/cvpr/HeZRS16}, we also consider the VGG-16~\cite{DBLP:conf/iclr/SimonyanZ14a} network where the \texttt{fc7} layer is used as the visual feature vector. In Table~\ref{tab:local}, we can observe that the ResNet-50 network with the \texttt{block3} feature achieves the best performance on all three datasets. Hence we choose to use the \texttt{block3} layer by default. Moreover, we find using VGG features is generally worse than using ResNet features.

\begin{table}[t]
\footnotesize
\centering
\begin{tabularx}{\linewidth}{l *3{>{\centering\arraybackslash}X}@{}}
% \toprule
\textbf{Local Feature}														&	\textbf{Fashion-1}	&	\textbf{Fashion-2}	&	\textbf{Home}\\	\hline
\textbf{ResNet-50}\\
\texttt{block1} (28$\times$28$\times$256)												&	68.0				&	74.5			&	77.7	 \\
\texttt{block2} (14$\times$14$\times$512)												&	67.6				&	73.3			& 	78.3	\\
\texttt{block3} (7$\times$7$\times$1024) 												&	\textbf{68.5}	&	\textbf{75.3}	&	\textbf{79.6}	 \\
\texttt{block4} (7$\times$7$\times$2048)												&	64.9				&	74.2			&	78.5	 \\
\midrule
\textbf{VGG-16}\\
% pool1												&	00.8	&		&	00.6	 \\
% pool2												&	00.1	&	.1	& .5	\\
\texttt{pool3} (28$\times$28$\times$512)									&	64.0	&	71.1	&	76.2	 \\
\texttt{pool4} (14$\times$14$\times$512)									&	63.3	&	71.7	&	76.7	 \\
\texttt{pool5} (7$\times$7$\times$512)										&	63.1	&	72.0	&	75.5	 \\
% \bottomrule
\end{tabularx}
\caption{The effect of local features}
\label{tab:local}

\end{table}

\subsection{Performance on Each Category}

\begin{figure}[t]
\includegraphics[align=t,width=.49\linewidth]{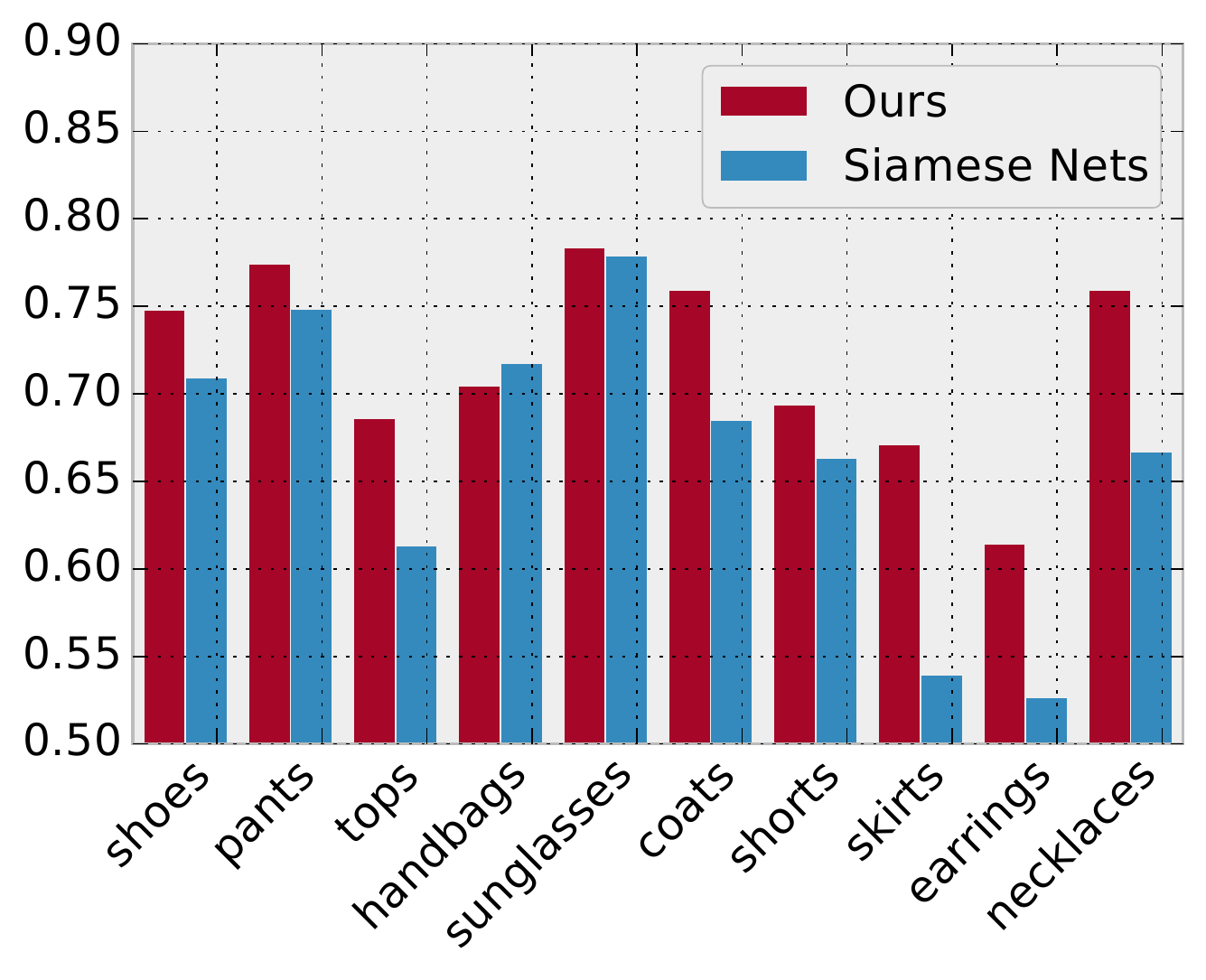}
\includegraphics[align=t,width=.49\linewidth]{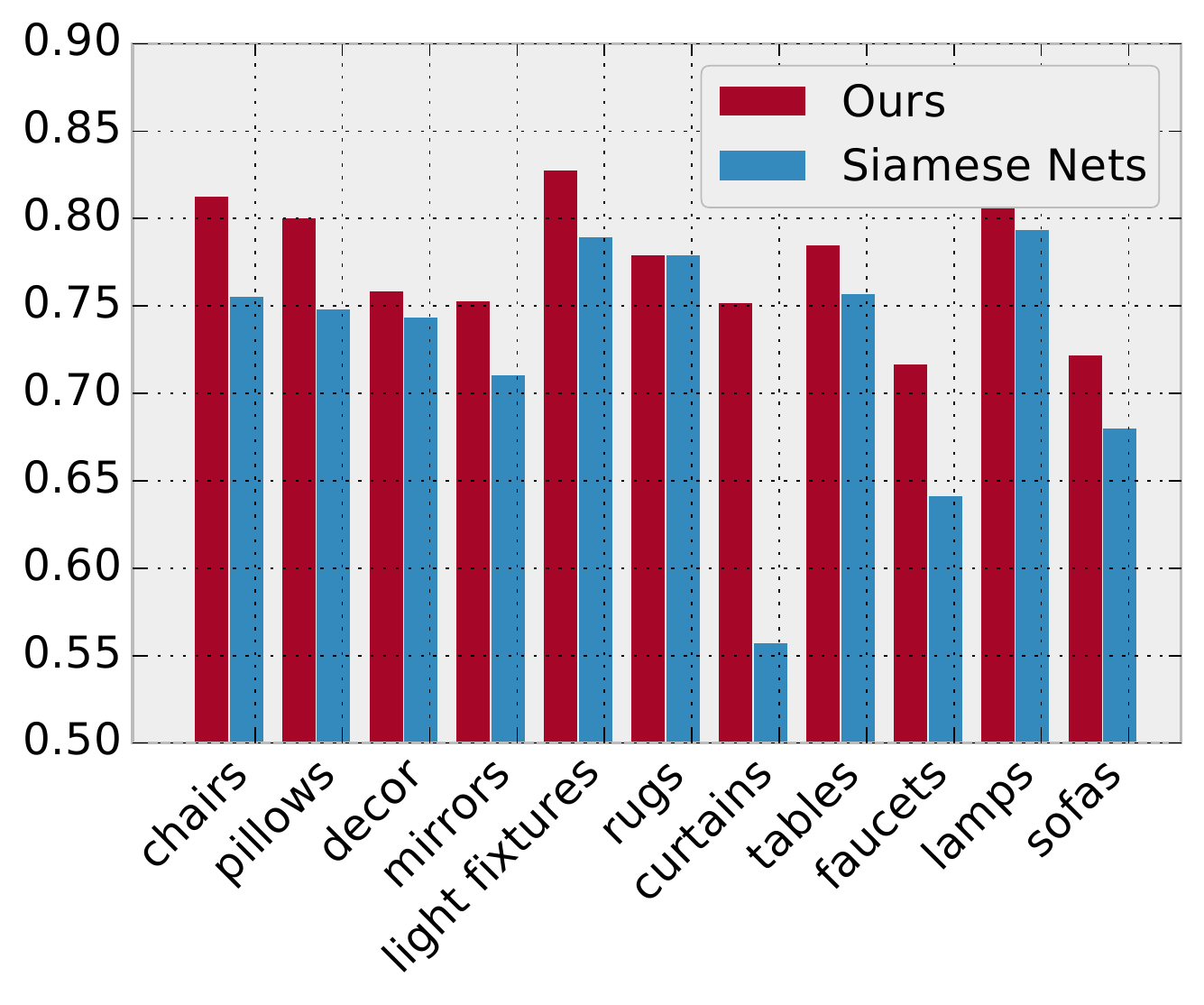}
\caption{Accuracy per category. Left: STL-Fashion, right: STL-Home.}
\label{fig:cat}
\end{figure}

We exam the performance on each category, compared with the strongest baseline Siamese Nets~\cite{DBLP:conf/iccv/VeitKBMBB15}. Figure~\ref{fig:cat} shows the performance comparison per category, based on the STL-Fashion and STL-Home datasets. We can see that in most categories, our method outperforms Siamese Nets. Moreover, the baseline has severe performance drops on categories like skirts and curtains. This verifies the effectiveness of our method when recommending products from various categories.
\section{Implementation Details}

The architecture of the two-layer network $g(\Theta;\cdot)$ is \texttt{Linear-BN-Relu-Dropout-Linear-L2Norm}, where the dimensionality is set to 4$\times d$ for the first linear layer, and $d$ for the last linear layer. The dropout rate is set to 0.5, and the learning rate is set to 0.001. For all methods, we update the statistics of batch normalization~\cite{DBLP:conf/icml/IoffeS15} layers in ResNet during training, and find it generally improves the performance.

\section{Qualitative Examples}

\begin{figure}[t]
    \centering
    \includegraphics[width=\linewidth]{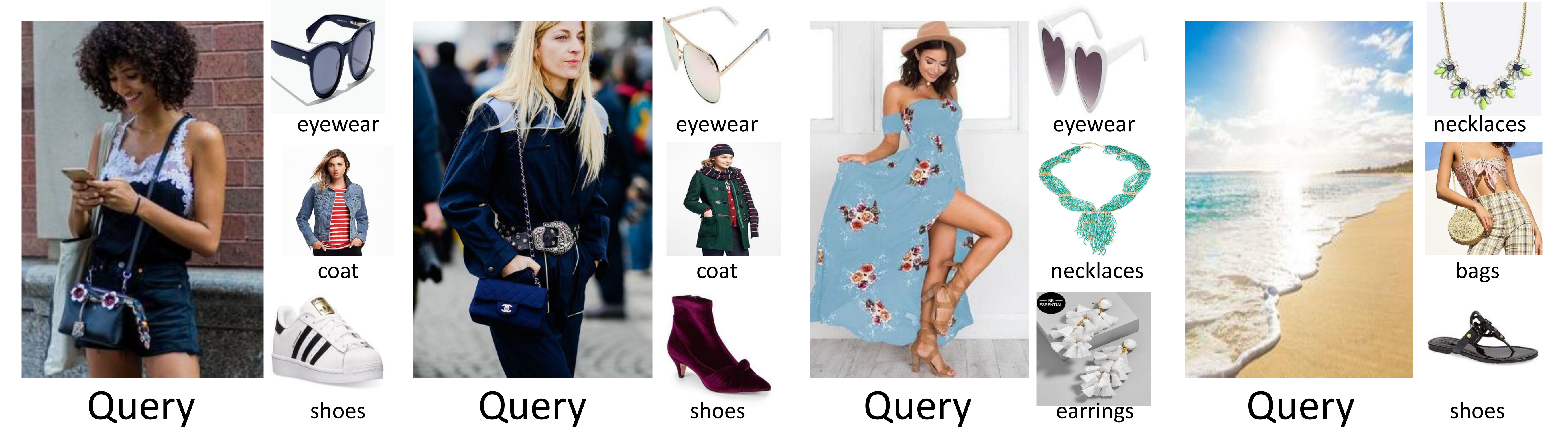}
    \caption{Recommended items with scenes \emph{outside} of our datasets. Left image is query, right composes top recommended products.}
    \label{fig:my_label}
\end{figure}

In addition to the examples where the query scenes are from the test set (as shown in the paper), we also test the model with scenes outside of the dataset~(Figure~\ref{fig:my_label}). 
We obtained them as top search results for queries `street fashion',`women fashion', and `beach' on Google. For each query, we show top-1 products from three categories.

\section{Human Study Interface}

Figure~\ref{fig:human} shows screenshots of the interface when conducting the user study. We first provide a description of the task and a small sample of tests to the fashion experts, and then ask them to answer the questions (i.e., choose one of the two products that is more compatible with the given scene).

{\small
\bibliographystyle{ieee}
\bibliography{egbib}
}

\begin{figure*}
\centering
\includegraphics[height=.3\textheight,width=0.8\linewidth]{figs/human1}
\includegraphics[height=.3\textheight,width=0.8\linewidth]{figs/human3}
\includegraphics[height=.3\textheight,width=0.8\linewidth]{figs/human2}
\caption{Screenshots of the user study interface.}
\label{fig:human}
\end{figure*}